\documentclass[10pt,twocolumn,letterpaper]{article}

\usepackage[pagenumbers]{cvpr} 

\definecolor{cvprblue}{rgb}{0.21,0.49,0.74}
\usepackage[pagebackref,breaklinks,colorlinks,allcolors=cvprblue]{hyperref}
\usepackage[accsupp]{axessibility}
\usepackage[margin=20mm]{geometry}
\usepackage{makecell}
\usepackage{diagbox}

\def\paperID{32} 
\def\confName{CVPRW}
\def\confYear{2026}

\title{IA-CLAHE: Image-Adaptive Clip Limit Estimation for CLAHE}

\author{
Rikuto Otsuka \and
Yuho Shoji \and
Yuka Ogino \and  
Takahiro Toizumi \and
Atsushi Ito \and
NEC Corporation \\
{
\tt\small \{otsuka-rikuto,yuho-shoji,yogino,t-toizumi\_ct,ito-atsushi\}@nec.com}
}

\begin{document}
\maketitle

\begin{abstract}
This paper proposes image-adaptive contrast limited adaptive histogram equalization (IA-CLAHE). Conventional CLAHE is widely used to boost the performance of various computer vision tasks and to improve visual quality for human perception in practical industrial applications. CLAHE applies contrast limited histogram equalization to each local region to enhance local contrast. However, CLAHE often leads to over-enhancement, because the contrast-limiting parameter clip limit is fixed regardless of the histogram distribution of each local region. Our IA-CLAHE addresses this limitation by adaptively estimating tile-wise clip limits from the input image. To achieve this, we train a lightweight clip limits estimator with a differentiable extension of CLAHE, enabling end-to-end optimization. Unlike prior learning-based CLAHE methods, IA-CLAHE does not require pre-searched ground-truth clip limits or task-specific datasets, because it learns to map input image histograms toward a domain-invariant uniform distribution, enabling zero-shot generalization across diverse conditions. Experimental results show that IA-CLAHE consistently improves recognition performance, while simultaneously enhancing visual quality for human perception, without requiring any task-specific training data.
\end{abstract}

\begin{figure}[t]
  \centering
   \includegraphics[width=1.0\linewidth]{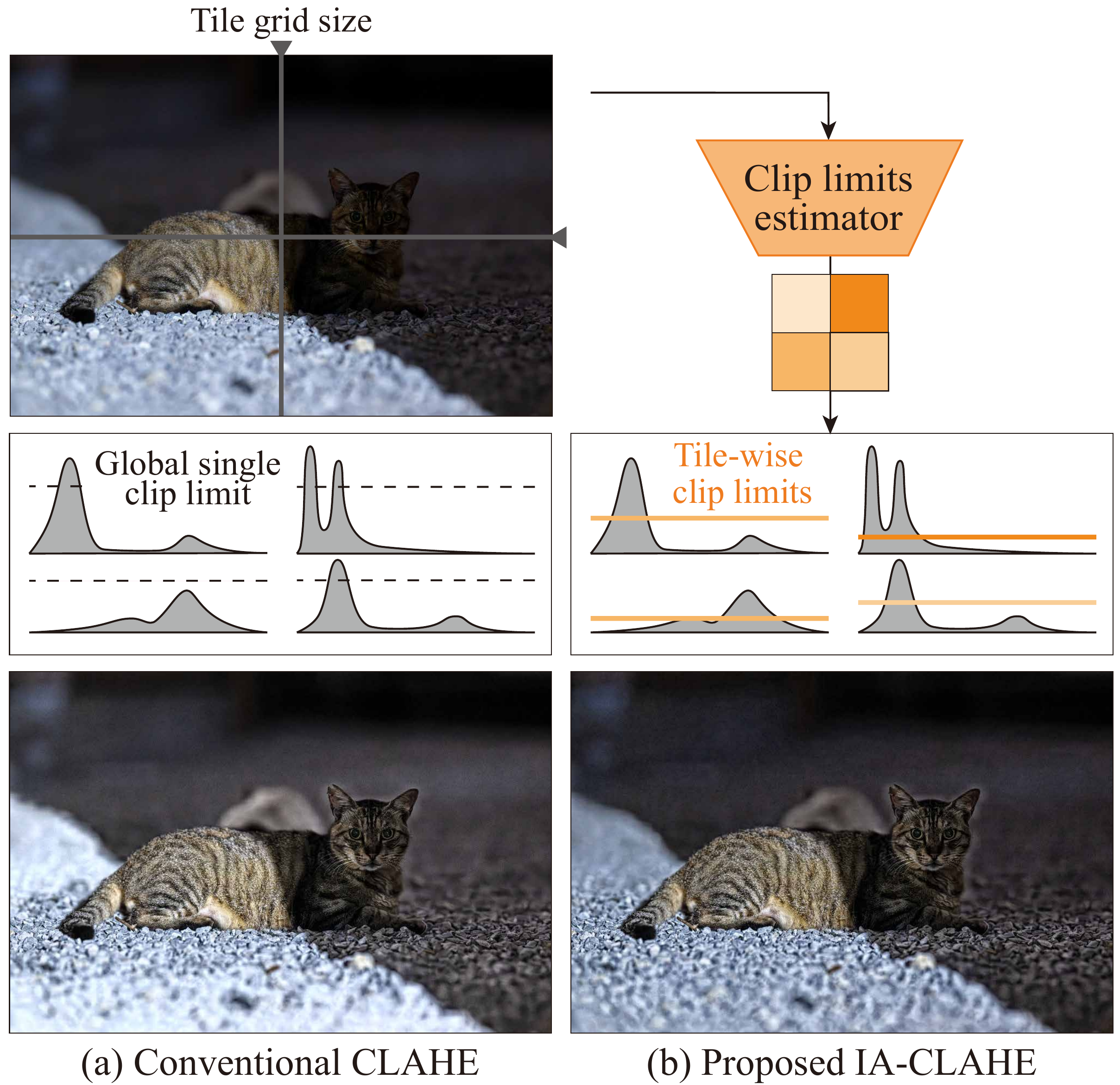}
   \caption{
   Comparison between conventional CLAHE and the proposed IA-CLAHE.
   (a) Conventional CLAHE applies a single global clip limit to all tiles, visualized as the dashed line in the middle-left histogram. This fixed parameter often causes over-enhancement in already bright regions.
   (b) IA-CLAHE predicts tile-wise clip limits (orange lines in the middle-right histogram) using a lightweight estimator and applies them through a differentiable CLAHE module. This adaptive strategy enhances details in dark regions while preserving well-exposed areas.
   }
   \label{fig:hdrcat}
\end{figure}    
\section{Introduction}
\label{sec:intro}
Images captured under adverse weather conditions such as low-light or dense fog often suffer from reduced contrast, amplified noise, and weakened structural information, which obscure object boundaries and degrade both human visibility and machine perception. These degradations significantly reduce the performance of downstream computer vision tasks, including classification and object detection.

To address these issues, a wide range of image enhancement techniques has been developed. Physical-model-based approaches \cite{he2010_single,liu2021_retinex} leverage assumptions about scene radiance and transmission, but their applicability is limited to specific degradation types (e.g., fog models fail under nighttime conditions). End-to-end restoration methods \cite{jiang2021_enlightengan,sun2024_restoring} based on U-Net-like architectures achieve strong performance on trained conditions but generalize poorly to unseen degradations. In contrast, tone-mapping-based methods \cite{guo2020_zero,li2022_zerodcepp,Hui2022_IA3DLUT} directly manipulate pixel intensities via parametric mappings or look-up tables (LUTs), offering high computational efficiency suitable for real-time applications.

In recent years, tone-mapping-based methods have been explored for improving machine recognition rather than human visibility \cite{liu2022_image,xu2023_toward,ogino2025_erup, otsuka2025_rethinking}. Thanks to their low computational cost, they can be seamlessly integrated into recognition pipelines as a real-time pre-processing stage, demonstrating high adaptability across various downstream tasks. Although effective, these methods typically require task-specific datasets of adverse weather images, whose collection and annotation are costly. To overcome this limitation and achieve domain generalization, histogram-based methods \cite{Parletta2023_AdaptiveHE,otsuka2025_rethinking} have been proposed. These approaches transforms the histogram of the input image to a desired distribution regardless of the histogram of input image. Therefore, the approaches can enhance the visibility of the image of unseen domain in zero-shot scenario.

Among various histogram-based methods, we focus on contrast limited adaptive histogram equalization (CLAHE) \cite{zuiderveld1994_contrast}. CLAHE has been widely used due to its local adaptivity, noise suppression capability, and computational efficiency. CLAHE divides an input image into non-overlapping tiles and applies histogram equalization locally while limiting the maximum bin count with a clip limit. However, its performance is highly sensitive to the choice of the clip limit, and a fixed global value often leads to over-enhancement depending on local histogram characteristics.

Several prior works have attempted to determine optimal clip limits using handcrafted rules \cite{Kamel2023_IQRCLAHE} or search-based optimization \cite{min2013_novel,more2015parameter,fawzi2021_adaptive,narla2024_multimodal,Wang2024_IVEFCLAHE,han2025_bo}, but these approaches either show limited generalization or require high computational cost. Learning-based CLAHE methods \cite{Liu2017_espcclahe,Campos19_clahe,Li2024_vggclahe} estimate a clip limits using machine learning models, yet they are restricted to a single global clip limit because the histogram clipping and redistribution steps in CLAHE have been considered non-differentiable, preventing end-to-end optimization. 
As a result, existing methods rely on a two-stage pipeline—exhaustive search for ground-truth clip limits followed by regression—which is computationally expensive during training and lacks the flexibility to scale across different datasets, loss functions, or adapt to data augmentation.

In this work, we revisit the formulation of CLAHE and show that its histogram redistribution process is almost everywhere differentiable with respect to the clip limits. This key differentiability enables us to design image-adaptive CLAHE (IA-CLAHE), an end-to-end trainable framework that estimates tile-wise clip limits using a lightweight CNN and applies them through a differentiable CLAHE module. This differentiability is essential for gradient-based optimization of the CNN parameters, because the CLAHE processing is connected after the neural network to be trained, followed by the loss calculation. 
As illustrated in Figure \ref{fig:hdrcat}, IA-CLAHE adaptively adjusts the tile-wise clip limits for each tile, mitigating the over-enhancement issues inherent in conventional CLAHE with a fixed global parameter.

Our main contributions are summarized as follows:

\begin{itemize}
    \item We show that CLAHE is almost everywhere differentiable with respect to clip limit parameters, enabling end-to-end optimization without ground-truth clip limit values.
    \item We propose IA-CLAHE, a lightweight framework that estimates tile-wise clip limits and improves the visibility of input images without causing over-enhancement.
    \item We demonstrate that IA-CLAHE generalizes in a zero-shot manner, improving both visual quality and recognition performance in adverse weather conditions without task-specific training dataset.
\end{itemize}
\begin{figure}[t]
  \centering
  \begin{subfigure}{\linewidth}
    \centering
    \includegraphics[width=\linewidth]{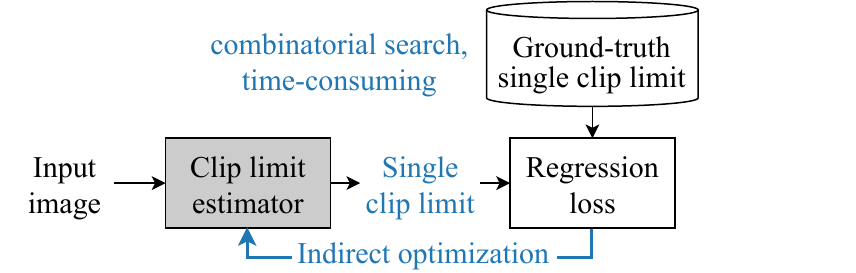}
    \caption{Conventional learning-based CLAHE methods}
    \label{fig:teaser:a}
  \end{subfigure}
  \vspace{2mm}
  \begin{subfigure}{\linewidth}
    \centering
    \includegraphics[width=\linewidth]{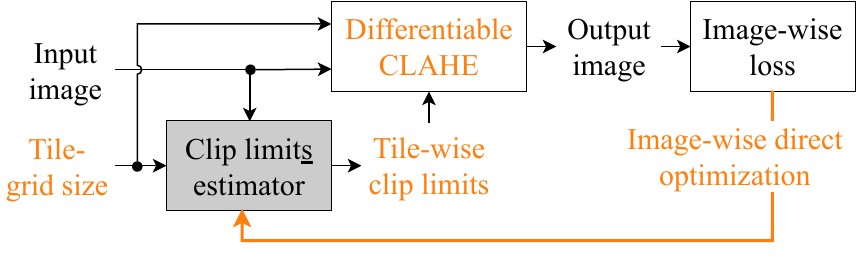}
    \caption{Proposed image-adaptive CLAHE (IA-CLAHE)}
    \label{fig:teaser:b}
  \end{subfigure}
  \caption{(a) The conventional learning-based CLAHE methods requires pre-searched ground-truth clip limit values to train a clip limit estimator via a regression loss, which involves combinatorial search and is time-consuming. (b) The proposed image-adaptive CLAHE can train a clip limits estimator directly from the input image via image-wise end-to-end optimization through differentiable CLAHE, without requiring ground-truth clip limit values.}
  \label{fig:teaser}
\end{figure}

\section{Related work}
\subsection{Tone-Mapping-Based Image Enhancement}
Tone-mapping-based image enhancement methods aim to improve visibility by transforming pixel intensities or histogram distributions through parametric mappings. Representative approaches include ZeroDCE \cite{guo2020_zero} and ZeroDCE++ \cite{li2022_zerodcepp}, which estimates pixel-wise tone curves, and IA-3DLUT \cite{Hui2022_IA3DLUT}, which blends multiple trained LUTs based on the input image. These methods are computationally efficient and suitable for real-time applications, with numerous extensions proposed in recent years \cite{Canqian2022a_AdaInt,Fengyi2022_CLUT,Ziwen2024_Real}.

Although tone-mapping-based methods have been applied to both human visibility enhancement and machine recognition \cite{liu2022_image,xu2023_toward,ogino2025_erup}, they generally require training on datasets under adverse weather condition, which limits their scalability. In contrast, CLAHE provides strong local adaptivity and noise suppression without task-specific training, motivating our focus on CLAHE-based enhancement.

\subsection{Histogram Equalization and CLAHE}
\label{sec:HEandCLAHE}
Global Histogram Equalization (HE) \cite{rafael2022_intensity} enhances contrast by matching the image histogram to a uniform distribution, but it ignores local contrast and often produces over-enhanced or visually unnatural results. Contrast limited adaptive histogram equalization (CLAHE) \cite{pizer1990_contrast,zuiderveld1994_contrast} addresses this by dividing an image into tiles and applying local histogram equalization while clipping histogram bin counts above a specified clip limit. This design improves local adaptivity and suppresses noise amplification, which contributes to the use of CLAHE widely in practical applications. Rule-based methods derive clip limits from handcrafted statistics such as entropy \cite{Anis2015_isoclahe}, percentile thresholds \cite{Kamel2023_IQRCLAHE}, or local intensity variance \cite{Chang2018_gammaclahe}. While simple, these heuristics often fail to generalize across diverse scenes. Search-based methods optimize clip limits using image quality metrics such as PSNR or entropy through heuristic search \cite{min2013_novel,kumari2025qualitative} or meta-heuristics \cite{more2015parameter,narla2024_multimodal,han2025_bo, kumari2025qualitative}. However, their computational cost grows rapidly with the number of tiles, making tile-wise optimization impractical.

Learning-based CLAHE methods \cite{Liu2017_espcclahe,Campos19_clahe,Li2024_vggclahe} train machine learning models to estimate a clip limit directly from an input image. For example, VGG‑16 \cite{simonyan2014_very} or XGBoost regressor \cite{Chen2016_xgboost} have been used to estimate a clip limit. However, these methods are restricted to a single global clip limit, because the histogram clipping and redistribution in CLAHE have traditionally been regarded as non-differentiable, preventing end-to-end training. Consequently, as illustrated in Figure \ref{fig:teaser:a}, learning-based CLAHE methods rely on a two-stage pipeline: searching for ground-truth clip limits and then regressing them. 
This search-then-regress approach becomes particularly challenging when extending to tile-wise clip limits, as the search time grows exponentially with the number of tiles. For a $T_H \times T_W$ tiles with $N$ candidate values per tile, exhaustive search requires $O(N^{T_{H}T_{W}})$ CLAHE evaluations, which is impractical. Even when clip limits are searched for each tile independently, $O(NT_{H}T_{W})$ CLAHE evaluations per image are still needed. This independent search remains computationally expensive\footnote{For instance, with $T_H = T_W = 8$ and $N = 100$, processing a single image takes 2.13 seconds on an RTX 3080 GPU, requiring 4.27 days for ground-truth pre-searching across our training dataset.}. This is computationally expensive during training and lacks the flexibility to scale across different datasets, loss functions, or adapt to data augmentation.

\section{Differentiability Analysis of CLAHE}
\label{sec:clahe}

\begin{figure}[t]
    \centering
    \includegraphics[width=1.00\linewidth]{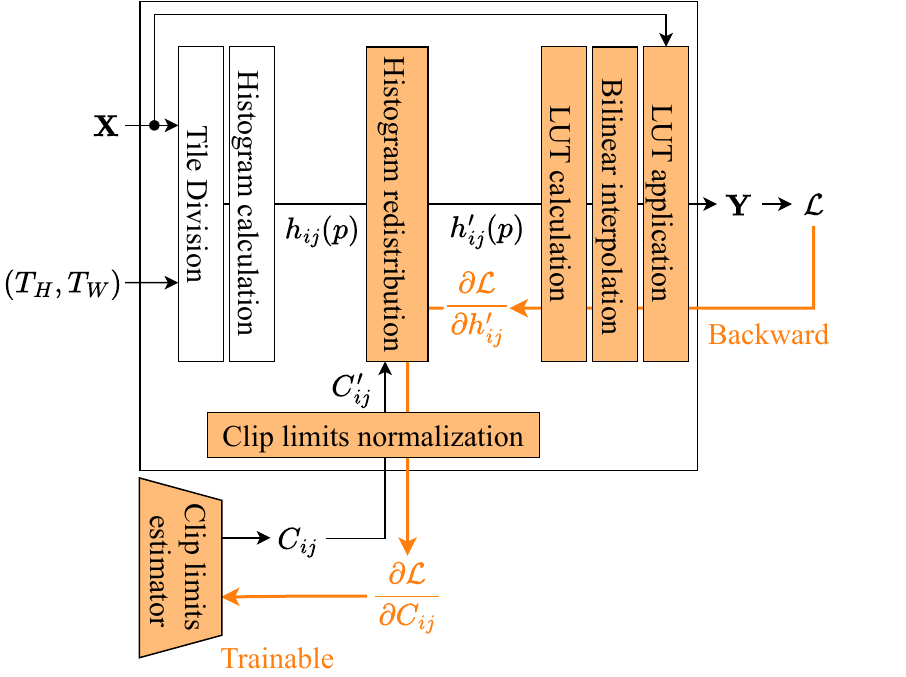}
    \caption{The overall pipeline of CLAHE. Among the CLAHE pipeline components, histogram redistribution involves discrete bin count redistribution, making its differentiability non-trivial. Therefore, we derive analytical expression for the gradient of the histogram redistribution operation with respect to the clip limit parameters. Other components, including LUT calculation via CDF computation, bilinear interpolation, and LUT application, are either linear or inherently differentiable, and are thus omitted here.}
    \label{fig:clahe_pipeline}
\end{figure}

This section revisits contrast limited adaptive histogram equalization (CLAHE) \cite{zuiderveld1994_contrast} and show that CLAHE is almost everywhere differentiable with respect to tile-wise clip limits, enabling direct optimization with image-wise L1 loss without ground-truth clip limit values.

\subsection{Overall Pipeline of CLAHE}
Figure \ref{fig:clahe_pipeline} illustrates the overall pipeline of CLAHE. CLAHE takes image $\mathbf{X} \in \mathbb{R}^{H \times W}$, tile grid size $(T_H, T_W)$, and tile-wise clip limits $\mathbf{C}=(C_{ij})\in\mathbb{R}^{T_H\times T_W}$ with $i\in\{1,\dots,T_H\}$ and $j\in\{1,\dots,T_W\}$ as input. Note that in practice, a clip limit $C_{ij}=C$ for all tiles is typically used to avoid the complexity of tile-wise parameter tuning.

CLAHE first divides an input image $\mathbf{X}$ into non-overlapping $T_H \times T_W$ tiles. For each tile $t_{ij}$, CLAHE then computes a local histogram $h_{ij}(p)$, which counts the frequency of pixel intensities $p \in \mathcal{P} = \{0, \ldots, P\}$, where $P$ is the maximum intensity (e.g., $P=255$ for 8-bit images).

After histogram calculation, CLAHE normalizes tile-wise clip limits for histogram redistribution. Normalized tile-wise clip limits $C_{ij}'$ are calculated as follows:
\begin{equation}
    C'_{ij} = C_{ij}N_{tile}/N_{bin},
\end{equation}
where $N_{bin}$ denotes the number of histogram bins, and $N_{tile} = T_H \times T_W$ denotes the total number of tiles. 

By using the normalized tile-wise clip limits $C'_{ij}$, the histogram $h_{ij}(p)$ is clipped and redistributed: histogram bin counts exceeding $C'_{ij}$ are clipped to $C'_{ij}$, and the total excess count \(S_{ij} = \sum_{p\in \mathcal{P}} \max\!\left(0,\, h_{ij}(p) - C'_{ij}\right)\) is redistributed uniformly across all bins. The redistributed histogram $h'_{ij}(p)$ is calculated as follows:
\begin{equation}
h'_{ij}(p) =
\begin{cases}
    C'_{ij}, & C'_{ij} \leq h_{ij}(p),\\
    h_{ij}(p) + S_{ij}/N_{bin}, & h_{ij}(p) < C'_{ij}.
\end{cases}
\end{equation}

Using the redistributed histogram $h'_{ij}(p)$, CLAHE calculates tile-wise look-up tables (LUTs) for each tile $t_{ij}$. It calculates a cumulative distribution function over intensity levels from $h'_{ij}(p)$. 
\begin{equation}
\label{eq:lut}
LUT_{ij}(p) = P\,\frac{\sum_{q=0}^{p} h'_{ij}(q)}{T_H\,T_W}.
\end{equation}

CLAHE then upsamples the tile-wise LUTs using bilinear interpolation based on the center coordinates of each tile. This upsampling produces a pixel-wise LUT for each pixel location $(x,y)$ to ensure smooth transformations between tile regions. Finally, CLAHE applies pixel-wise LUT to the input image $\mathbf{X}$ to generate the output image $\mathbf{Y}$ by replacing each pixel value with its corresponding mapped value.

\subsection{Differentiability with Respect to Clip Limits}
\label{sec:clahe:diff}
Our goal is to backpropagate gradients from arbitrary loss (e.g. image reconstruction loss and other task-specific losses) to the tile-wise clip limit parameters $C_{ij}$. As highlighted in orange in the Figure \ref{fig:clahe_pipeline}, our method needs to establish differentiability from histogram redistribution through to the loss computation. The LUT calculations and bilinear interpolation have been addressed in prior works \cite{Hui2022_IA3DLUT, Canqian2022a_AdaInt} and are known to be differentiable. We focus on deriving the gradients with respect to the clip limit values during the histogram redistribution step.

The gradient of loss $\mathcal{L}$ with respect to clip limit parameter $C_{ij}$ is obtained by the chain rule:
\begin{equation}
\frac{\partial \mathcal{L}}{\partial C_{ij}}
=
\frac{\partial C'_{ij}}{\partial C_{ij}}
\sum_{p\in \mathcal{P}}
\frac{\partial h'_{ij}(p)}{\partial C'_{ij}}
\frac{\partial \mathcal{L}}{\partial h'_{ij}(p)},
\label{eq:gradient}
\end{equation}
where $\frac{\partial \mathcal{L}}{\partial h'_{ij}(p)}$ represents the gradient backpropagated from the subsequent differentiable operations in CLAHE (CDF computation, interpolation, and the final image reconstruction). The partial derivatives in Equation \ref{eq:gradient} are derived as:
\begin{align}
    \frac{\partial C_{ij}'}{\partial C_{ij}} &= \frac{N_{tile}}{N_{bin}},\label{eq:grad_C_a}\\
    \frac{\partial h_{ij}'(p)}{\partial C_{ij}'} &=
    \begin{cases}
        1, \quad C'_{ij} \leq h_{ij}(p)\\
       - N_{bin}^{-1} \sum_{q \in \mathcal{P}} \mathbf{1}\!\left(h_{ij}(q) > C'_{ij}\right), \quad\text{otherwise} &
    \end{cases}
    \label{eq:grad_C_b}
\end{align}
where, $\mathbf{1}(condition)$ denotes the indicator function that returns $1$ if the $condition$ is satisfied and $0$ otherwise. By combining these gradients, we can train a model to estimate optimal tile-wise clip limits $C_{ij}$ by arbitrary loss functions.
\section{Proposed Method}
\label{sec:ProposedMethod}
The proposed IA-CLAHE is an end-to-end trainable framework that estimates tile-wise clip limits. We first introduce the overall pipeline. Figure \ref{fig:teaser:b} shows the framework of IA-CLAHE. Given an input image $\mathbf{X} \in \mathbb{R}^{H \times W}$ and tile grid size $(T_H, T_W)$, a lightweight clip limits estimator estimates a tile-wise clip limits $\mathbf{C}=(C_{ij})\in\mathbb{R}^{T_H\times T_W}$ with $i\in\{1,\dots,T_H\}$ and $j\in\{1,\dots,T_W\}$. The estimated tile-wise clip limits $C_{ij}$ is fed into a differentiable CLAHE  (orange area in Figure \ref{fig:clahe_pipeline}) to generate an output image $\mathbf{Y}$. During training, an image-wise L1 loss $\mathcal{L}$ is computed between the output $\mathbf{Y}$ and the ground truth image, and parameters of the clip limits estimator are optimized via backpropagation.

\subsection{Clip Limits Estimator}
\label{sec:ClipLimitsEstimator}
\begin{figure}[t]
  \centering
   \includegraphics[width=\linewidth]{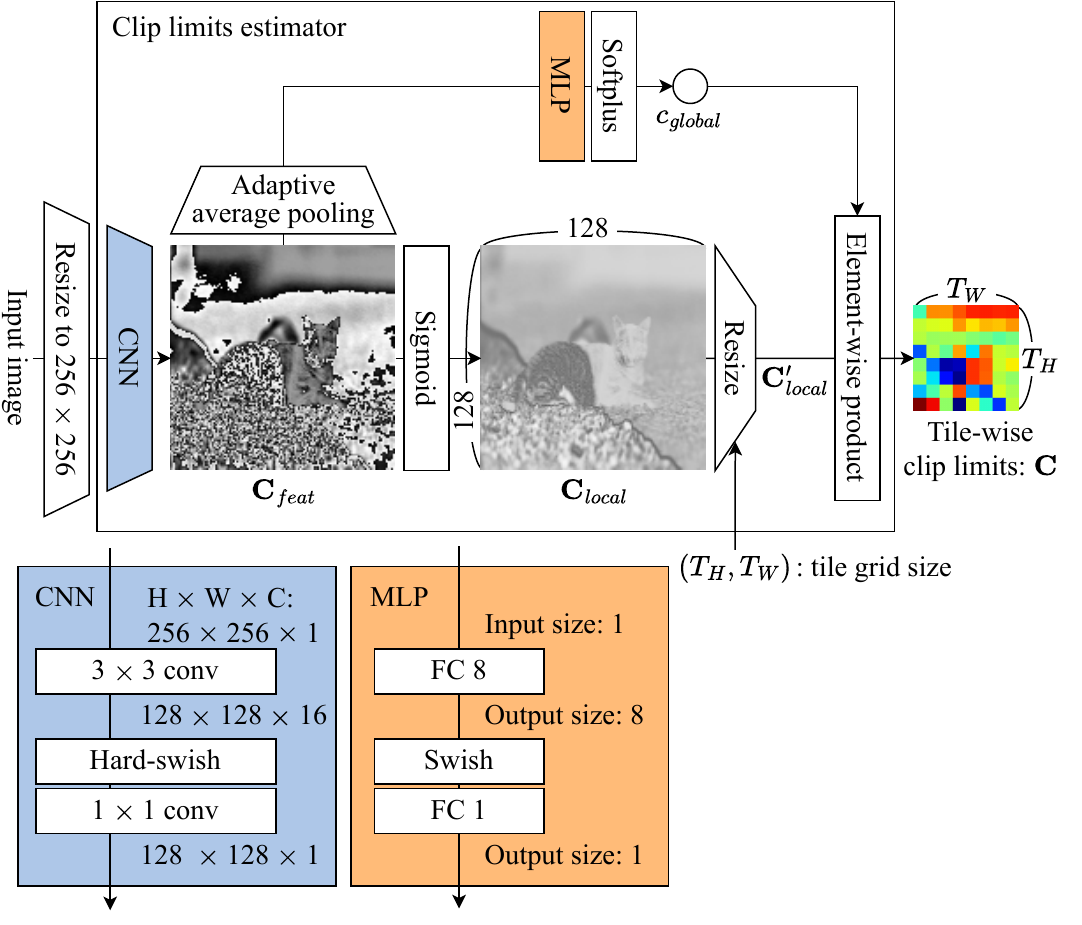}
   \caption{Detailed architecture of the clip limits estimator. This design enables clip limits estimation with a limited number of parameters, making the estimator suitable for real-time deployment.}
   \label{fig:cnn_mlp}
\end{figure}

We design a clip limit estimator using a lightweight CNN to estimate the tile-wise clip limits \(\mathbf{C}_{ij}\) from the input image, conditioned on the tile grid size. Figure \ref{fig:cnn_mlp} illustrates the detailed architecture of the clip limits estimator. We extract a single Y channel of YCbCr color space from the input image and resize it from \(H \times W\) to \(256 \times 256\) before feeding it into the estimator. The estimator first extracts features $\mathbf{C}_{feat}$ from the resized image, and then obtains a local map $\mathbf{C}_{local}$ and a global scale factor $c_{global}$. The local map determines which regions are enhanced and the map can be resized to an arbitrary tile-grid size. The global scale factor is a scalar value that adjusts the value range of the clip limits.

For feature extraction, we use a CNN block. The CNN block includes a $3\times3$ convolutional layer (stride = 2, padding = 1 and bias disabled), a hard-swish activation, and $1\times1$ convolutional layer (bias disabled) layers. We initialize the $3\times3$ convolutional layer using weights of the first convolution layer of the ImageNet-pretrained MobileNetV3 \cite{howard2019_searching}. Because these pre-trained weights are designed for 3-channel RGB input, we convert the RGB weights into YCbCr color space and use the weight corresponding to the Y channel. The CNN block extracts a one-channel feature map $\mathbf{C}_{feat}$ with the size of $128 \times 128$. 

For local map estimation, We apply a sigmoid function on the $\mathbf{C}_{feat}$ and obtain the local map $\mathbf{C}_{local}$. This local map is then resized to the tile grid size $(T_H, T_W)$ using bilinear interpolation, and obtain $\mathbf{C}'_{local}$. For global scale factor estimation, we apply adaptive average pooling to $\mathbf{C}_{feat}$ and extract a $1\times 1$ global feature. This value is fed into an MLP consisting of two fully connected layers with a swish activation. MLP output is activated by softplus function to ensure positive value. Finally, the element-wise product between $c_{global}$ and $\mathbf{C}'_{local}$ provides the clip limits $\mathbf{C}$. 

We trained the clip limits estimator using L1 loss between clean and degraded images. During inference, we specify $(T_H, T_W)$ as a hyperparameter and resize $\mathbf{C}_{local}'$. To enable this flexible adaptation to arbitrary tile grid sizes, we randomly sample $(T_H, T_W)$ during training, which also prevents the estimated clip limits from converging to uniform values across all tiles. The predicted local map $\mathbf{C}_{local}$ can be rescaled to different tile grid sizes without retraining and enables flexible adaptation to various scene conditions.
\section{Experiments}
We evaluated our IA-CLAHE on image recognition and image enhancement tasks in a zero-shot scenario. We first assessed the performance on image classification and object detection tasks using three datasets collected under adverse weather conditions. We then evaluated the image enhancement performance on two low-visibility datasets. We additionally demonstrate that our IA-CLAHE maintains a runtime comparable to that of conventional CLAHE.

\subsection{Implementation Details of IA-CLAHE}
\label{sec:trainingIACLAHE}
We convert the RGB image to the YCbCr color space \cite{2011_ycbcr}, extract the Y channel, and use it as input to IA-CLAHE. We evaluated our IA-CLAHE with tile grid sizes of $(T_H \times  T_W) = (1 \times 1)$ and $(8 \times 8)$. Under the $(1 \times 1)$ configuration, the method becomes histogram equalization with a single global clip limit $C$ applied to the entire image. For region-wise processing, the tile grid size is set to $(8 \times 8)$, which is common in prior works \cite{min2013_novel,Parletta2023_AdaptiveHE} and implemented as the default setting in OpenCV \cite{Bradski2000_opencv}. A comparison between the $(1 \times 1)$ and $(8 \times 8)$ settings demonstrates that the suitable grid size depends on the task and confirms the advantage of our method in enabling flexible tile grid size tuning.

We used the Adam optimizer \cite{Kingma14_adam} to train our clip limits estimator. The learning rate was set to $1.0\times 10^{-4}$ for $17,680$ iterations with a batch size of $1$. For training, we used clean images from the MSEC dataset \cite{Afifi21_msec} to fairly evaluate downstream task performance in a zero-shot scenario. This dataset is constructed from the MIT-Adobe FiveK dataset \cite{Bychkovsky11_adobe5k}. Following prior work \cite{Afifi21_msec}, we used images retouched by expert photographer C as clean input images. The input image size was set to $640 \times 640$ pixels. For data augmentation, we applied two histogram manipulation techniques used in \cite{Campos19_clahe}: histogram compression and intensity shift. Histogram compression scales the dynamic range of the histogram by a factor $\alpha$, while intensity shift translates the histogram using a constant value $\beta$. The parameters $\alpha$ and $\beta$ were randomly sampled from $[-0.5, 0.9]$ and $[-60, 60]$, respectively.

\subsection{Comparison Methods}
\label{sec:expRecog} 
We compared the proposed method with two tone-mapping-based methods, two recent learning-based methods, and three histogram-based methods. As tone-mapping-based methods, we used ZeroDCE++ \cite{li2022_zerodcepp}, IA-3DLUT \cite{Hui2022_IA3DLUT}. These methods estimate image-adaptive tone-mapping and enhance the contrast of an input image. We also compared with transformer \cite{sun2024_restoring} and diffusion-based \cite{wang2024_pqp} methods. As HE-based methods, we used HE \cite{rafael2022_intensity}, CLAHE \cite{zuiderveld1994_contrast}, rule-based CLAHE method (RB-CLAHE \cite{Kamel2023_IQRCLAHE}), and learning-based CLAHE methods (LB-CLAHE \cite{Campos19_clahe}). We follow the parameter settings of their original paper to implement these methods. All methods are trained using only normal-light images in order to evaluate zero-shot generalization.

\begin{table}[t]
    \small
    \centering
    \caption{Comparison results on image classification on CODaN and object detection on ExDark and DAWN. The best result for each column is highlighted in {\bf bold}.}
    \label{tab:lowlight}
    \begin{tabular}{c|cc|c|c}
        \hline
        & \multicolumn{2}{c|}{CODaN} & ExDark & DAWN \\
        \hline
        \diagbox{Methods}{Condition} & Day & Night & Night & \makecell{Adverse\\weather}\\
        \hline
        w/o enhancement                                    & {\bf81.2} & 50.1 & 0.705 & 0.671\\
        ZeroDCE++ \cite{li2022_zerodcepp}                  & 78.7 & 58.9 & 0.702 & 0.601\\
        IA-3DLUT \cite{Hui2022_IA3DLUT}                    & 80.2 & 53.2 & 0.701 & 0.659\\
        Transformer-based \cite{sun2024_restoring}         & {\bf81.2} & 49.8 & 0.666 & 0.580\\
        Diffusion-based \cite{wang2024_pqp}                & 80.3 & 57.8 & 0.664 & 0.563 \\
        HE \cite{rafael2022_intensity}                     & 77.5 & 51.6 & 0.680 & 0.644\\
        CLAHE ($1\times 1$) \cite{zuiderveld1994_contrast} & 77.7 & 52.2 & 0.684 & 0.669 \\
        CLAHE ($8\times 8$) \cite{zuiderveld1994_contrast} & 66.8 & 47.1 & 0.682 & 0.670 \\
        RB-CLAHE ($1\times 1$) \cite{Kamel2023_IQRCLAHE}   & 81.1 & 50.1 & 0.704 & 0.670 \\
        RB-CLAHE ($8\times 8$) \cite{Kamel2023_IQRCLAHE}   & 72.9 & 54.5 & 0.709 & 0.679 \\
        LB-CLAHE \cite{Campos19_clahe}                     & 79.7 & 58.4 & 0.710 & 0.679 \\
        \hline
        IA-CLAHE ($1\times 1$) & 80.4 & {\bf60.3} & 0.709 & 0.674 \\
        IA-CLAHE ($8\times 8$) & 78.7 & 58.9 & {\bf0.711} & {\bf0.686} \\
        \hline
    \end{tabular}
\end{table}

\subsection{Evaluation Settings}
\label{subsec:evalSetting}
\paragraph{Settings of Image Classification:}
We evaluated image classification performance under adverse weather conditions using the CODaN \cite{Lengyel21_ciconv}. CODaN is a daytime and nighttime dataset with 10 object classes. CODaN provides 10,000 daytime training images and 500 daytime validation images, as well as 3,000 daytime and 3,000 nighttime test images. Following prior work \cite{Lengyel21_ciconv}, we used a ResNet-18 \cite{He16_resnet} classifier trained only on the daytime training split. All images are resized to $224 \times 224$ pixels. We reported top-1 accuracy.

\begin{table*}[t]
    \centering
    \small
    \caption{Visual quality evaluation on the MSEC and LCDP datasets. The metrics are measured on the Y channel in the YCbCr color space. PSNR/SSIM indicate structural preservation, while BRISQUE/NIQE evaluate perceptual quality and contrast naturalness. An improvement in BRISQUE/NIQE together with a drop in PSNR/SSIM often suggests over-enhancement.}
    \label{tab:visual}
    \begin{tabular}{c|cccc|cccc}
        \hline
        Method & \multicolumn{4}{c|}{MSEC} & \multicolumn{4}{c}{LCDP} \\
        \hline
        & PSNR $\uparrow$ & SSIM $\uparrow$ & BRISQUE $\downarrow$ & NIQE $\downarrow$ & PSNR $\uparrow$ & SSIM $\uparrow$ & BRISQUE $\downarrow$ & NIQE $\downarrow$ \\
        \hline
        ZeroDCE++ \cite{li2022_zerodcepp} & 12.13 & 0.53 & 33.59 & 3.88 & 18.20 & 0.75 & 17.15 & \underline{3.41} \\
        IA-3DLUT \cite{Hui2022_IA3DLUT} & {\bf20.58} & {\bf0.85} & 29.68 & 3.57 & 16.55 & 0.56 & 28.29 & 3.89 \\
        Transformer-based \cite{sun2024_restoring} & 17.06 & 0.77 & 51.01 & 5.82 & 15.41 & 0.54 & 65.07 & 6.81\\
        Diffusion-based \cite{wang2024_pqp} & 18.04 & 0.76 & 41.51 & 5.59 & {\bf19.19} & \underline{0.74} & 48.30 & 6.32\\
        HE \cite{rafael2022_intensity} & 16.85 & 0.77 & 29.49 & 3.47 & 15.79 & 0.71 & 25.09 & 3.56 \\
        CLAHE ($1\times1$) \cite{zuiderveld1994_contrast} & 16.77 & 0.77 & 29.33 & 3.46 & 16.69 & 0.71 & 28.56 & 3.60 \\
        CLAHE ($8\times8$) \cite{zuiderveld1994_contrast} & 12.16 & 0.53 & 32.25 & {\bf3.22} & 13.28 & 0.51 & 40.07 & 3.60 \\
        RB-CLAHE ($1\times1$) \cite{Kamel2023_IQRCLAHE} & 17.23 & 0.80 & 29.71 & 3.72 & 15.47 & 0.60 & 20.56 & 3.80 \\
        RB-CLAHE ($8\times8$) \cite{Kamel2023_IQRCLAHE} & 17.64 & 0.81 & 27.58 & 3.52 & 15.59 & 0.60 & 19.94 & 3.76 \\
        LB-CLAHE \cite{Campos19_clahe} & 12.37 & 0.54 & 31.95 & \underline{3.23} & 14.05 & 0.53 & 39.06 & 3.61 \\
        \hline
        IA-CLAHE ($1\times1$)
        & \underline{20.08} & \underline{0.84} & \underline{27.52} & 3.44 & \underline{18.53} & {\bf0.76} & {\bf16.15} & 3.47 \\
        IA-CLAHE ($8\times8$)
        & 17.64 & 0.79 & {\bf26.04} & 3.27 & 17.77 & 0.72 & 17.35 & {\bf3.38} \\
        \hline
    \end{tabular}
\end{table*}

\begin{figure*}[t]
\centering
    \begin{subfigure}{0.19\linewidth}
    \centering
    \includegraphics
    [width=\linewidth,trim=0cm 212px 0cm 0cm,clip]
    {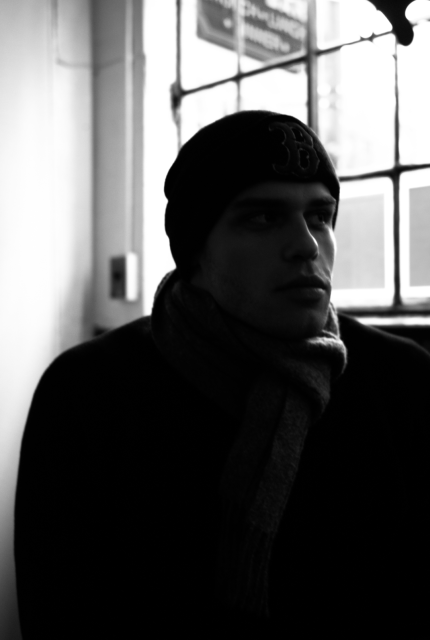}
    \caption*{Input}
    \end{subfigure}
    %
    %
    \begin{subfigure}{0.19\linewidth}
    \centering
    \includegraphics
    [width=\linewidth,trim=0cm 212px 0cm 0cm,clip]
    {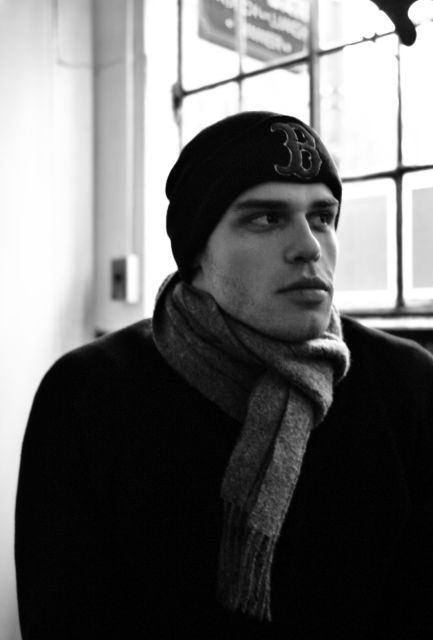}
    \caption*{ZeroDCE++ \cite{li2022_zerodcepp}}
    \end{subfigure}
    %
    %
    \begin{subfigure}{0.19\linewidth}
    \centering
    \includegraphics
    [width=\linewidth,trim=0cm 212px 0cm 0cm,clip]
    {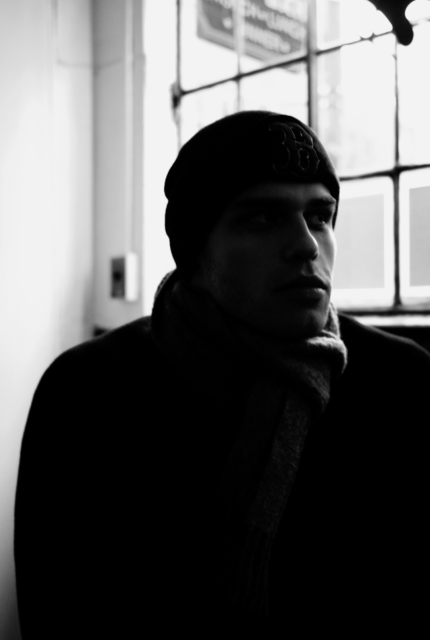}
    \caption*{IA-3DLUT \cite{Hui2022_IA3DLUT}}
    \end{subfigure}
    %
    %
    \begin{subfigure}{0.19\linewidth}
    \centering
    \includegraphics
    [width=\linewidth,trim=0cm 212px 0cm 0cm,clip]
    {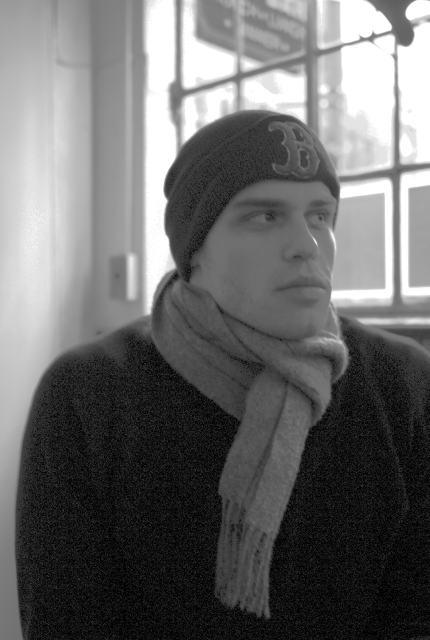}
    \caption*{HE \cite{rafael2022_intensity}}
    \end{subfigure}
    %
    %
    \begin{subfigure}{0.19\linewidth}
    \centering
    \includegraphics
    [width=\linewidth,trim=0cm 212px 0cm 0cm,clip]
    {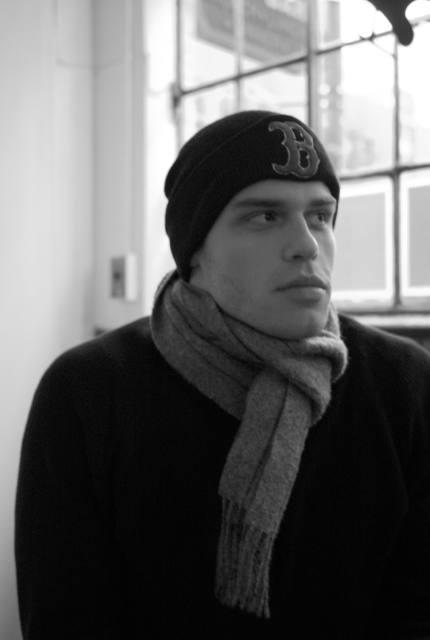}
    \caption*{Ground-truth}
    \end{subfigure}
    \vspace{1mm}
    %
    %
    \begin{subfigure}{0.19\linewidth}
    \centering
    \includegraphics
    [width=\linewidth,trim=0cm 212px 0cm 0cm,clip]
    {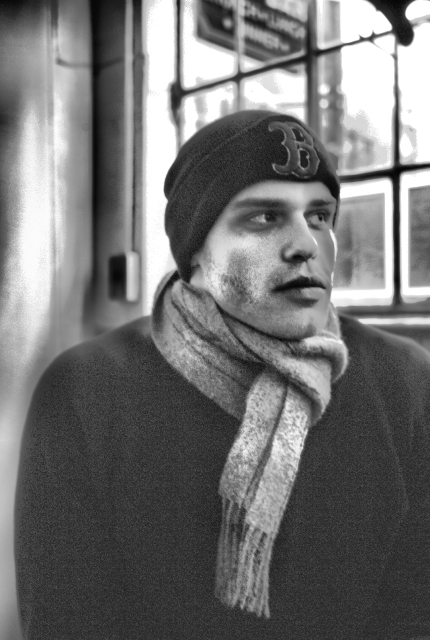}
    \caption*{CLAHE ($8\times 8$) \cite{zuiderveld1994_contrast}}
    \end{subfigure}
    %
    %
    \begin{subfigure}{0.19\linewidth}
    \centering
    \includegraphics
    [width=\linewidth,trim=0cm 212px 0cm 0cm,clip]
    {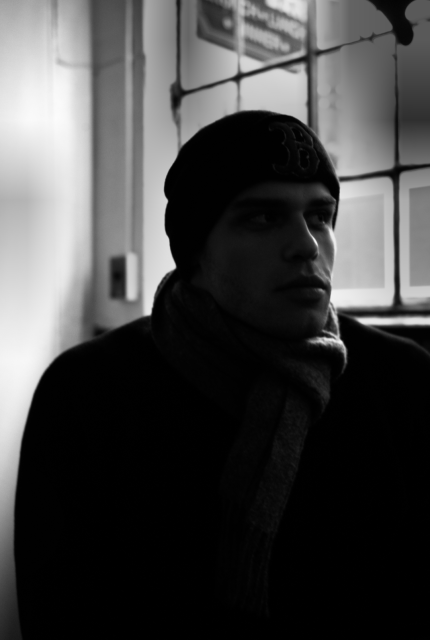}
    \caption*{RB-CLAHE ($8\times 8$) \cite{Kamel2023_IQRCLAHE}}
    \end{subfigure}
    %
    %
    \begin{subfigure}{0.19\linewidth}
    \centering
    \includegraphics
    [width=\linewidth,trim=0cm 212px 0cm 0cm,clip]
    {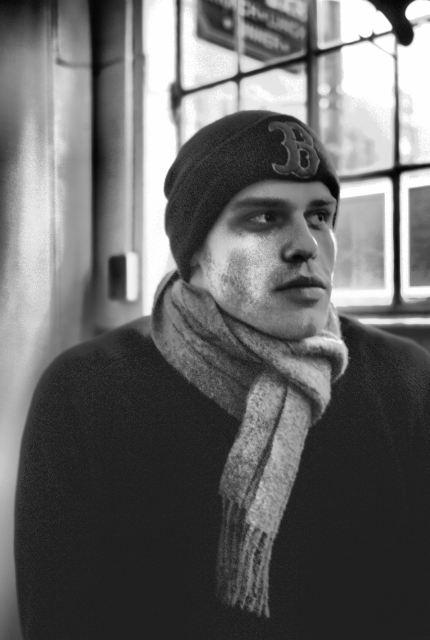}
    \caption*{LB-CLAHE \cite{Campos19_clahe}}
    \end{subfigure}
    %
    %
    \begin{subfigure}{0.19\linewidth}
    \centering
    \includegraphics
    [width=\linewidth,trim=0cm 212px 0cm 0cm,clip]
    {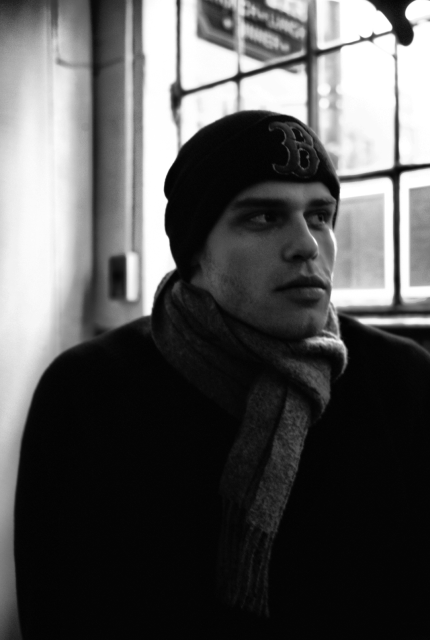}
    \caption*{IA-CLAHE ($8\times 8$)}
    \end{subfigure}
    %
    %
    \begin{subfigure}{0.19\linewidth}
    \centering
    \includegraphics
    [width=\linewidth]
    {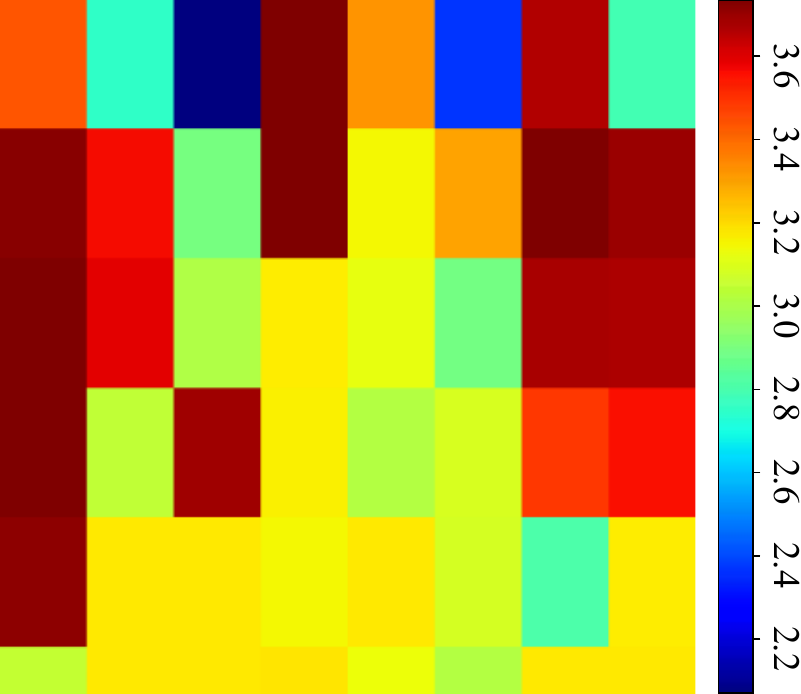}
    \caption*{Tile-wise clip limits}
    \end{subfigure}
\vspace{1mm}
%
%
    \begin{subfigure}{0.19\linewidth}
    \centering
    \includegraphics
    [width=\linewidth]
    {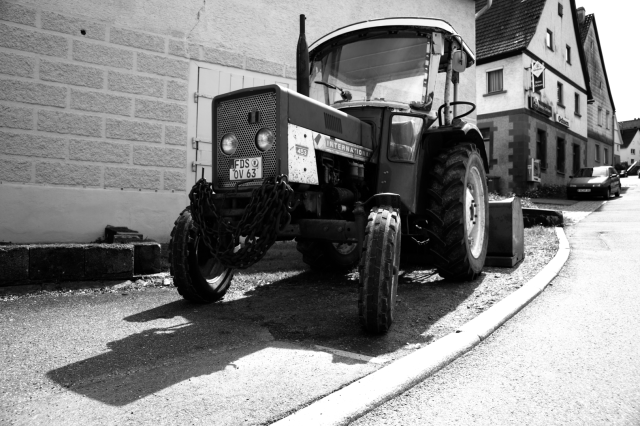}
    \vspace{1mm}
    \includegraphics
    [width=\linewidth,trim=5.5cm 3cm 0cm 7cm,clip]
    {imgs4publish/qualitative/a4748-kme_0969_input.png}
    \caption*{Input}
    \end{subfigure}
    %
    %
    \begin{subfigure}{0.19\linewidth}
    \centering
    \includegraphics
    [width=\linewidth]
    {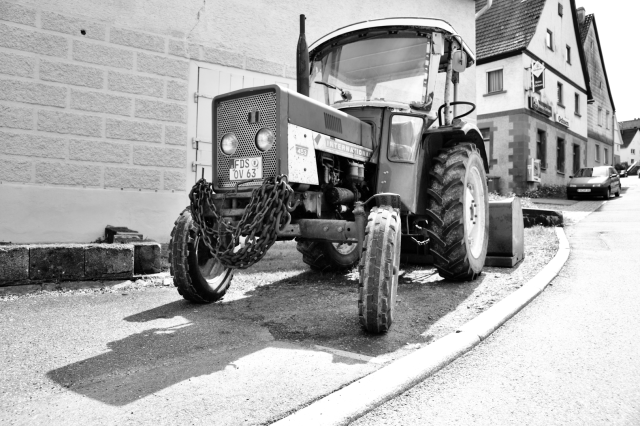}
    \vspace{1mm}
    \includegraphics
    [width=\linewidth,trim=5.5cm 3cm 0cm 7cm,clip]
    {imgs4publish/qualitative/a4748-kme_0969_zerodcepp.png}
    \caption*{ZeroDCE++ \cite{li2022_zerodcepp}}
    \end{subfigure}
    %
    %
    \begin{subfigure}{0.19\linewidth}
    \centering
    \includegraphics
    [width=\linewidth]
    {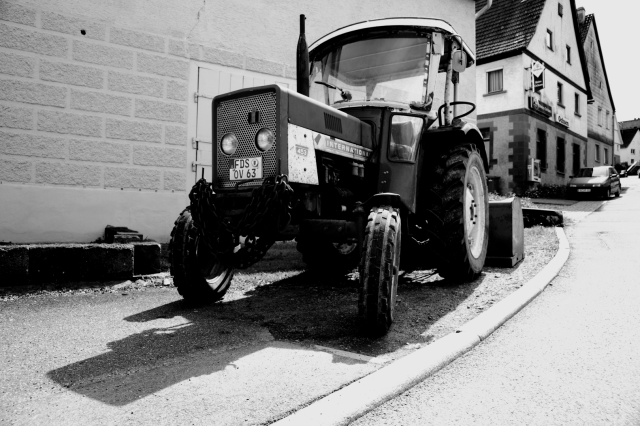}
    \vspace{1mm}
    \includegraphics
    [width=\linewidth,trim=5.5cm 3cm 0cm 7cm,clip]
    {imgs4publish/qualitative/a4748-kme_0969_ia3dlut.png}
    \caption*{IA-3DLUT \cite{Hui2022_IA3DLUT}}
    \end{subfigure}
    %
    %
    \begin{subfigure}{0.19\linewidth}
    \centering
    \includegraphics
    [width=\linewidth]
    {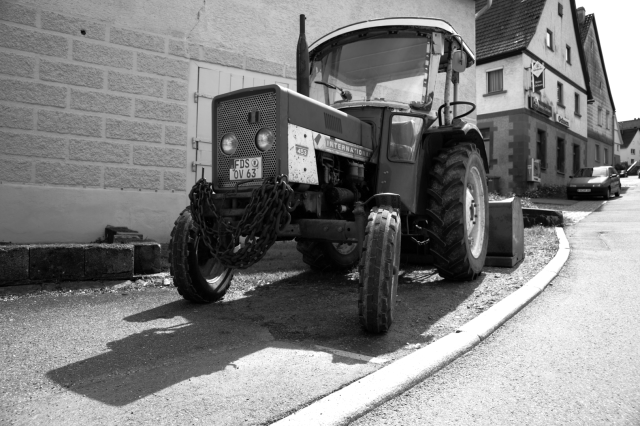}
    \vspace{1mm}
    \includegraphics
    [width=\linewidth,trim=5.5cm 3cm 0cm 7cm,clip]
    {imgs4publish/qualitative/a4748-kme_0969_he.png}
    \caption*{HE \cite{rafael2022_intensity}}
    \end{subfigure}
    %
    %
    \begin{subfigure}{0.19\linewidth}
    \centering
    \includegraphics
    [width=\linewidth]
    {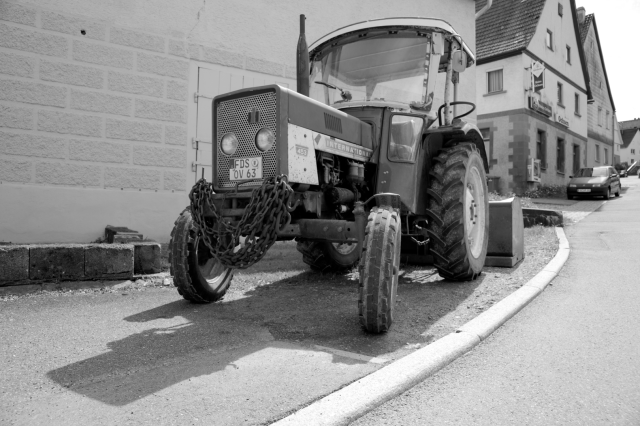}
    \vspace{1mm}
    \includegraphics
    [width=\linewidth,trim=5.5cm 3cm 0cm 7cm,clip]
    {imgs4publish/qualitative/a4748-kme_0969_gt.png}
    \caption*{Ground-truth}
    \end{subfigure}
\vspace{1mm}
%
%
    \begin{subfigure}{0.19\linewidth}
    \centering
    \includegraphics
    [width=\linewidth]
    {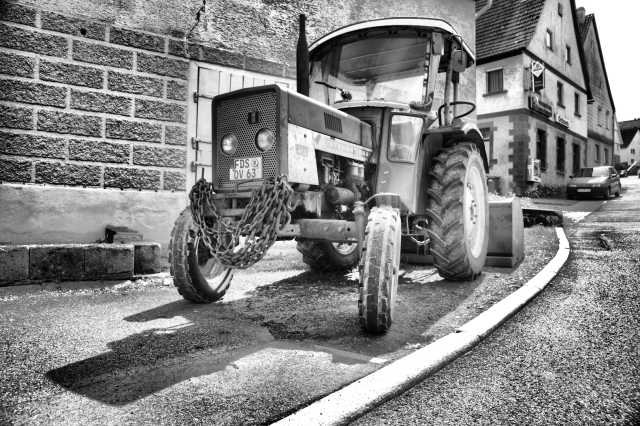}
    \vspace{1mm}
    \includegraphics
    [width=\linewidth,trim=5.5cm 3cm 0cm 7cm,clip]
    {imgs4publish/qualitative/a4748-kme_0969_clahe8840.png}
    \caption*{CLAHE ($8\times 8$) \cite{zuiderveld1994_contrast}}
    \end{subfigure}
    %
    %
    \begin{subfigure}{0.19\linewidth}
    \centering
    \includegraphics
    [width=\linewidth]
    {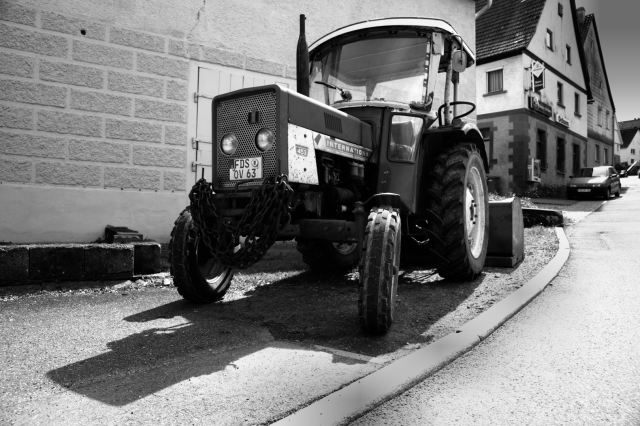}
    \vspace{1mm}
    \includegraphics
    [width=\linewidth,trim=5.5cm 3cm 0cm 7cm,clip]
    {imgs4publish/qualitative/a4748-kme_0969_rbclahe.png}
    \caption*{RB-CLAHE ($8\times 8$) \cite{Kamel2023_IQRCLAHE}}
    \end{subfigure}
    %
    %
    \begin{subfigure}{0.19\linewidth}
    \centering
    \includegraphics
    [width=\linewidth]
    {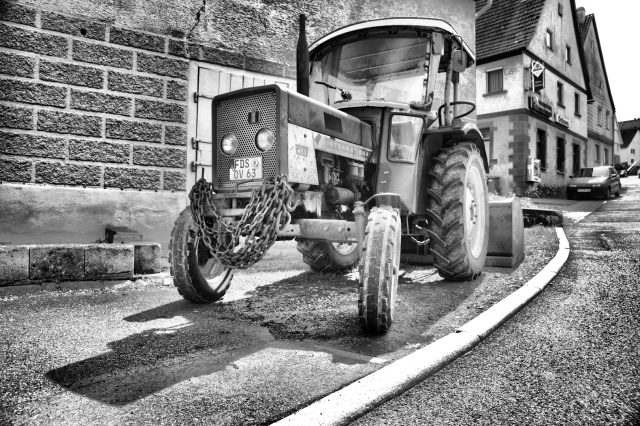}
    \vspace{1mm}
    \includegraphics
    [width=\linewidth,trim=5.5cm 3cm 0cm 7cm,clip]
    {imgs4publish/qualitative/a4748-kme_0969_lbclahe.png}
    \caption*{LB-CLAHE \cite{Campos19_clahe}}
    \end{subfigure}
    %
    %
    \begin{subfigure}{0.19\linewidth}
    \centering
    \includegraphics
    [width=\linewidth]
    {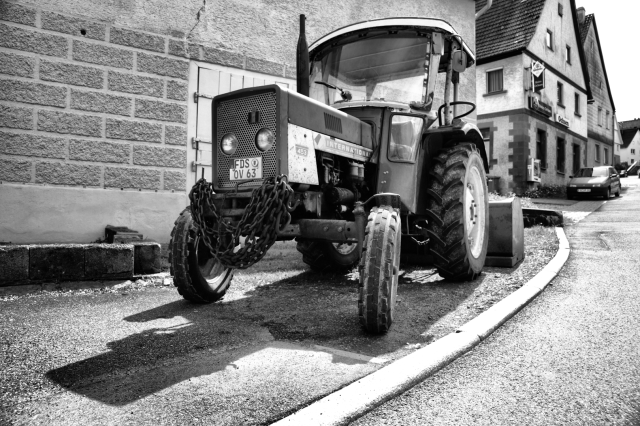}
    \vspace{1mm}
    \includegraphics
    [width=\linewidth,trim=5.5cm 3cm 0cm 7cm,clip]
    {imgs4publish/qualitative/a4748-kme_0969_iaclahe88.png}
    \caption*{IA-CLAHE ($8\times 8$)}
    \end{subfigure}
    %
    %
    \begin{subfigure}{0.19\linewidth}
    \centering
    \includegraphics
    [width=\linewidth]
    {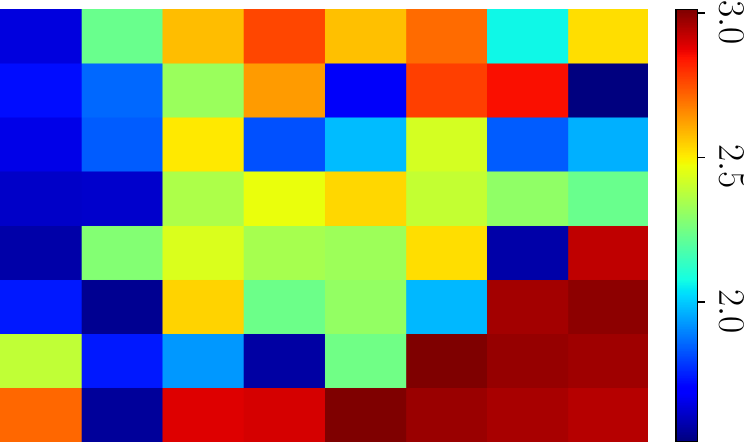}
    \vspace{1mm}
    \includegraphics
    [width=\linewidth,trim=5.5cm 3cm 0cm 7cm,clip]
    {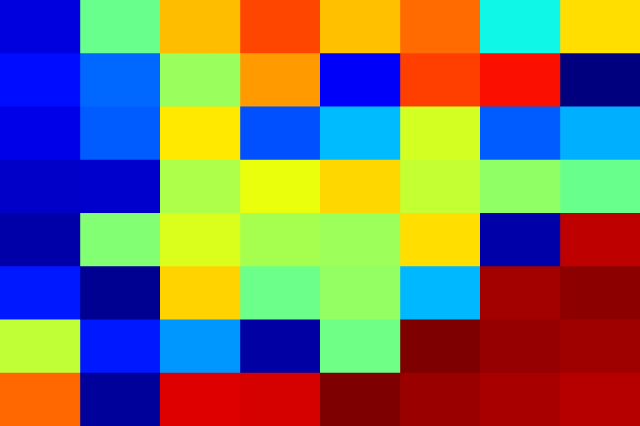}
    \caption*{Tile-wise clip limits}
    \end{subfigure}
\caption{Qualitative comparison on a LCDP dataset. IA-CLAHE assigns higher clip limits to low-contrast regions (e.g., windows, road on the right), effectively enhancing contrast in these challenging areas. In contrast, conventional CLAHE methods with single global clip limit tend to over-enhance already high-contrast regions (e.g., man's cheek, tractor), resulting in unnatural appearances.
}
\label{fig:qualitative_lcdp}
\end{figure*}

\paragraph{Settings of Object Detection:} We evaluate object detection performance under adverse weather conditions using ExDark \cite{Loh19_exdark} and DAWN \cite{Mourad2020_DAWN}. ExDark is a low-light dataset with 7,363 images across 12 object categories, each annotated with bounding boxes and class labels. DAWN is adverse weather dataset (fog, rain, sand, and snow) with 1,000 images across 6 object categories, each annotated with bounding boxes and class labels. As the detector, we use YOLOv3u \cite{redmon18_yolov3} from Ultralytics \cite{Glenn18_ultra}, pre-trained on MS COCO \cite{Lin14_mscoco}. Since MS COCO predominantly contains clear daytime images, this setup enables zero-shot evaluation under low-light (ExDark) and adverse weather conditions (DAWN). For both datasets, we set the input image size to $608 \times 608$ pixels and report mean average precision with a confidence threshold of $0.5$ and an intersection-over-union threshold of $0.45$, following prior work \cite{Li24_wiid}.

\paragraph{Settings of Visual Quality Evaluation}:
We evaluated IA-CLAHE on the test sets of the MSEC \cite{Afifi21_msec} and LCDP \cite{Wang2022_lcdp} datasets. The MSEC dataset contains 5,905 test images \cite{Ziwen2024_Real}. Following our training data, we used images retouched by photographer expert C as ground truth and evaluate enhancement quality using PSNR, SSIM \cite{Wang2004SSIM}, BRISQUE \cite{2012BRISQUE}, and NIQE \cite{2013NIQE}. The LCDP dataset \cite{Wang2022_lcdp} provides 218 test images with non-uniform illumination caused by simultaneous overexposure and underexposure. We reported both reference (PSNR/SSIM) and no-reference (BRISQUE/NIQE) metrics. PSNR/SSIM measure structural preservation (i.e., whether over-enhancement causes noise amplification and artifacts), while BRISQUE/NIQE measure perceptual quality and naturalness of contrast. This combination of reference and no-reference metrics are commonly used for evaluating CLAHE-based methods \cite{Campos19_clahe,Majeed_2020,han2025_bo,Majid_2025}. Since our approach primarily targets luminance enhancement, metrics are evaluated on the Y channel in YCbCr.

\subsection{Comparison Results on Image Recognition}
\label{subsec:ImageRecog}
Table \ref{tab:lowlight} reports top-1 accuracy on CODaN and mean average precision at an IoU threshold at 0.5 for object detection on ExDark and DAWN, where each enhancement method is applied as a preprocessing step to fixed pretrained recognition models. The preferred tile grid size differs between classification and object detection in this evaluation. A coarse grid (1$\times$1) is more effective for CODaN classification, whereas a finer grid (8$\times$8) is more effective for object detection on ExDark and DAWN. This is because images in classification dataset are cropped and often dominated by a single object. In contrast, object detection datasets contain multiple objects under spatially varying lighting, and processing each region enables contrast adjustment for each object.

Among CLAHE variants, learning-based CLAHE (LB-CLAHE, IA-CLAHE) consistently outperform rule-based CLAHE (RB-CLAHE), demonstrating stronger generalization for zero-shot image enhancement. Furthermore, IA-CLAHE achieves superior performance compared to LB-CLAHE, which estimates a single global clip limit. This suggests that the combination of tile-wise clip limits and direct optimization using image-wise L1 loss, both enabled by differentiable IA-CLAHE, contributes to the improved performance.
When compared with tone-mapping, transformer, and diffusion-based methods, our IA-CLAHE shows stronger robustness on night and adverse weather conditions. This is likely because CLAHE-based methods adaptively normalize local image histogram, reducing sensitivity to contrast variations under domain shifts.

\begin{table}[t]
  \small
  \centering
  \caption{Computation times [ms] and the number of parameters. For fairness, only PyTorch implementations were included. IA-CLAHE achieves nearly the same processing speed as CLAHE.}
  \label{tab:time}
  \begin{tabular}{c | r r | r}
    \hline
    Method & 4K& Full HD& \#Params \\
    \hline
    ZeroDCE++ \cite{li2022_zerodcepp} & 13.60 & 3.51 & 10,561 \\
    IA-3DLUT \cite{Hui2022_IA3DLUT}   & 2.48 & 0.86 & 593,516 \\
    Transformer-based \cite{sun2024_restoring} & 463 $<$ & 463 $<$ & $1.6\times 10^{7}$\\
    Diffision-based \cite{wang2024_pqp} & 4974 $<$ & 4974 $<$ & $1.3\times 10^{9}$\\
    CLAHE ($1\times1$) \cite{zuiderveld1994_contrast} & 2.19 & 0.92 & 3 \\
    CLAHE ($8\times8$) \cite{zuiderveld1994_contrast} & 2.17 & 0.89 & 3 \\
    RB-CLAHE ($1\times1$) \cite{Kamel2023_IQRCLAHE} & 2.19 & 0.92 & 3 \\
    RB-CLAHE ($8\times8$) \cite{Kamel2023_IQRCLAHE} & 2.17 & 0.89 & 3 \\
    \hline
    IA-CLAHE ($1\times1$) & 2.13 & 1.24 & 211\\
    IA-CLAHE ($8\times8$) & 2.95 & 0.89 & 211 \\
    \hline
  \end{tabular}
\end{table}

\begin{figure*}[t]
  \centering
  \begin{subfigure}[t]{0.19\linewidth}
    \centering
    \includegraphics[width=\linewidth]{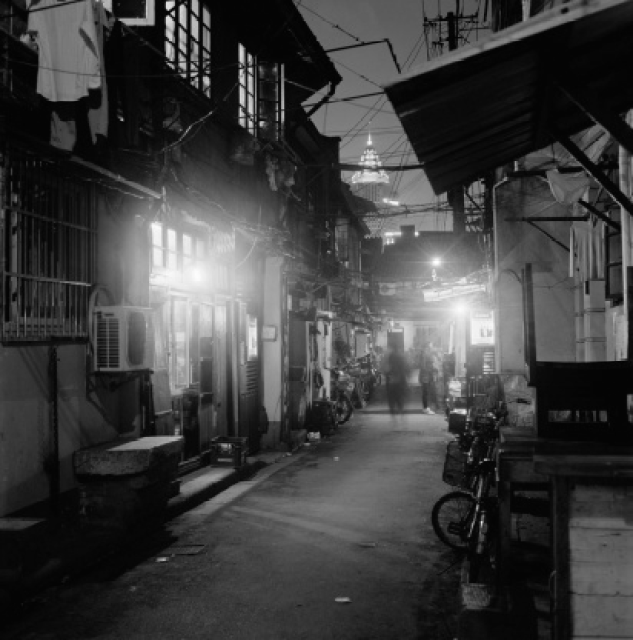}
    \caption{Input}
    \label{fig:flare:input}
    \vspace{-4mm}
  \end{subfigure}
  \begin{subfigure}[t]{0.19\linewidth}
    \centering
    \includegraphics[width=\linewidth]{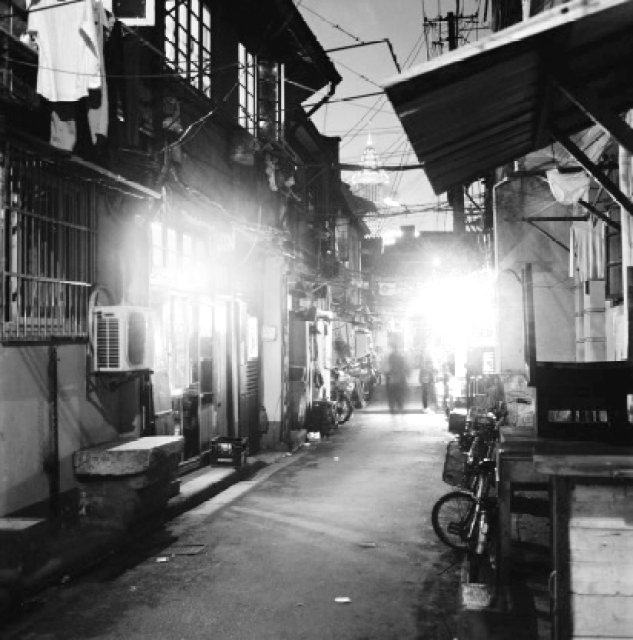}
  \end{subfigure}
  \begin{subfigure}[t]{0.19\linewidth}
    \centering
    \includegraphics[width=\linewidth]{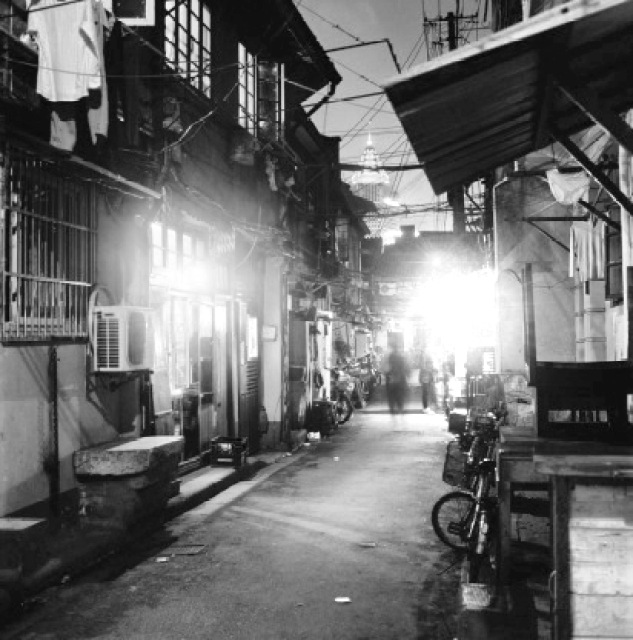}
  \end{subfigure}
  \begin{subfigure}[t]{0.19\linewidth}
    \centering
    \includegraphics[width=\linewidth]{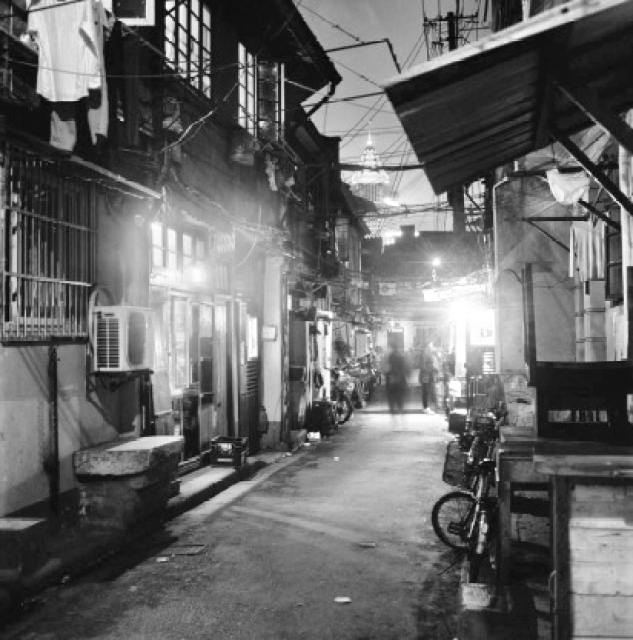}
  \end{subfigure}
  \begin{subfigure}[t]{0.19\linewidth}
    \centering
    \includegraphics[width=\linewidth]{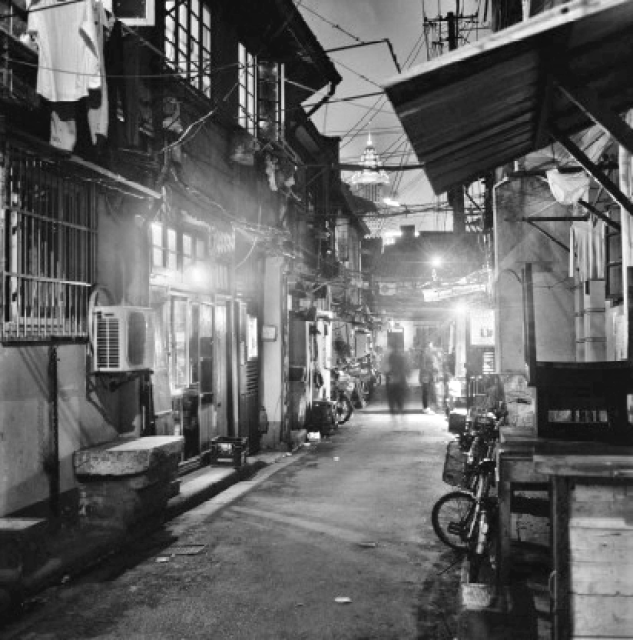}
  \end{subfigure}
  %
  %
  \begin{subfigure}[t]{0.19\linewidth}\centering
    \phantom{\includegraphics[width=\linewidth]{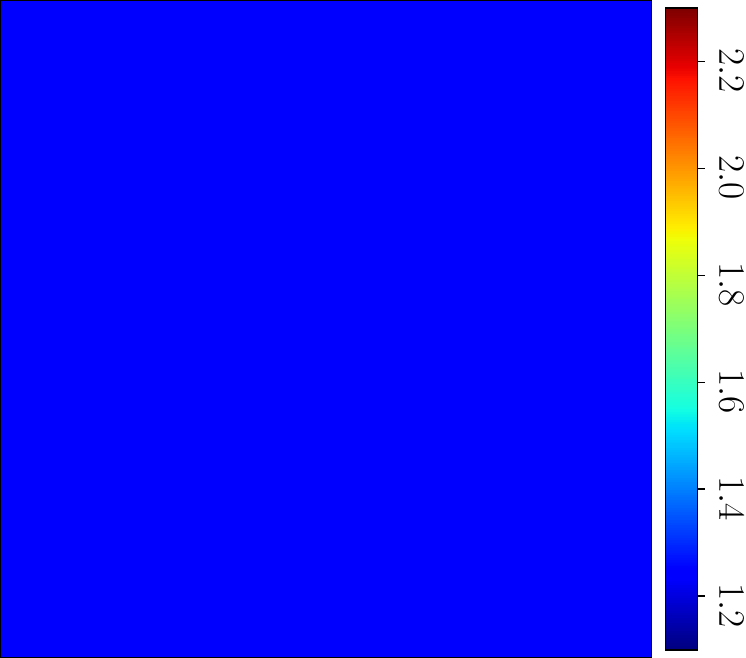}}
  \end{subfigure}
  \begin{subfigure}[t]{0.19\linewidth}
    \centering
    \includegraphics[width=\linewidth]{imgs4publish/ablation/cliplimit_heatmap/2015_00090_cliplimit_tgs1.pdf}
    \caption{IA-CLAHE ($1\times 1$)}
    \label{fig:flare:1x1}
  \end{subfigure}
  \begin{subfigure}[t]{0.19\linewidth}
    \centering
    \includegraphics[width=\linewidth]{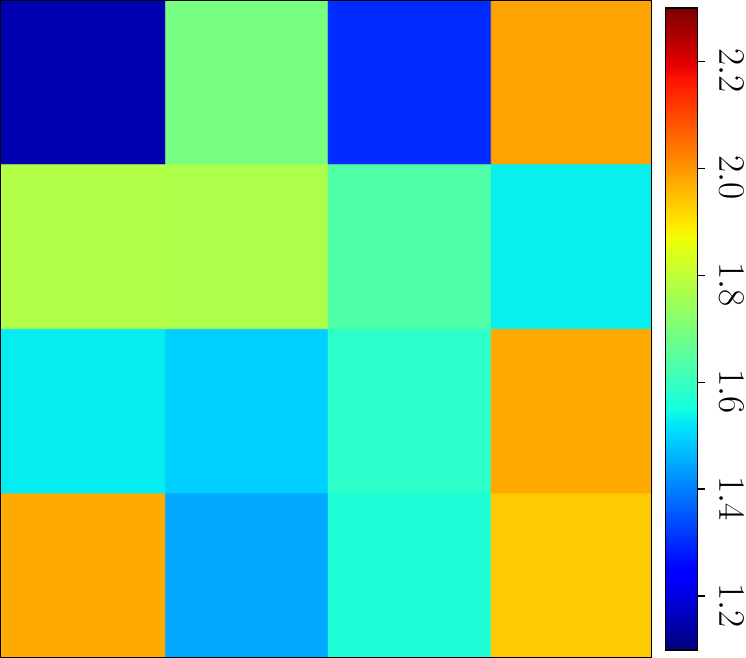}
    \caption{IA-CLAHE ($4\times 4$)}
    \label{fig:flare:4x4}
  \end{subfigure}
  \begin{subfigure}[t]{0.19\linewidth}
    \centering
    \includegraphics[width=\linewidth]{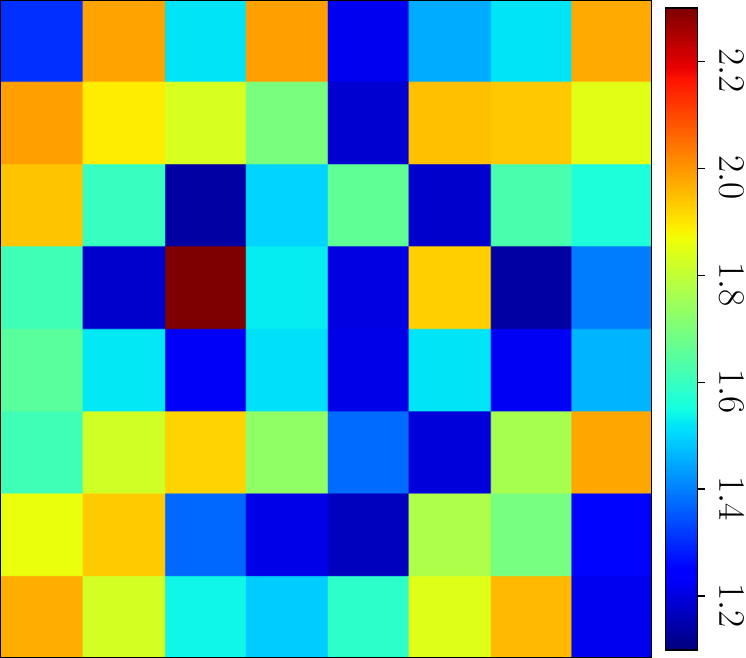}
    \caption{IA-CLAHE ($8\times 8$)}
    \label{fig:flare:8x8}
  \end{subfigure}
  \begin{subfigure}[t]{0.19\linewidth}
    \centering
    \includegraphics[width=\linewidth]{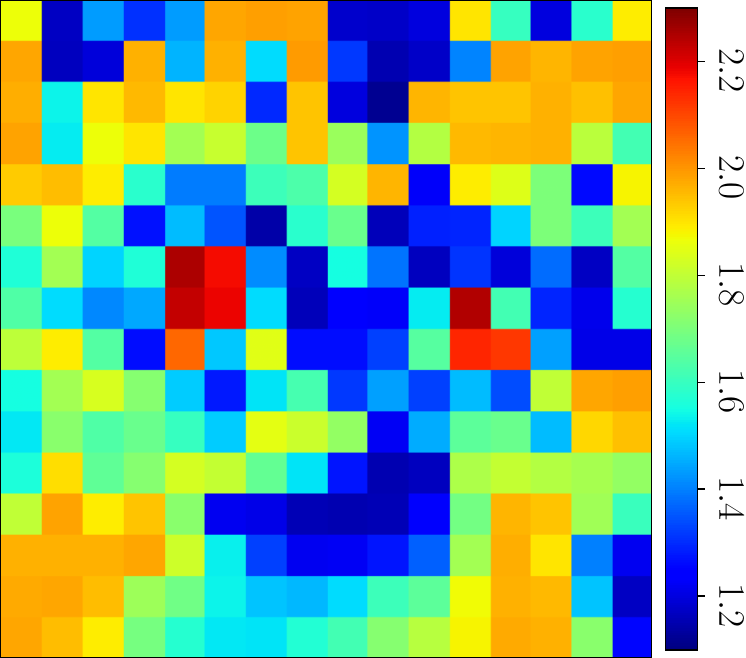}
    \caption{IA-CLAHE ($16\times 16$)}
    \label{fig:flare:16x16}
  \end{subfigure}
  %
  %
  %
  %
  %
  %
  %
  \begin{subfigure}[t]{0.19\linewidth}
    \centering
    \includegraphics[width=\linewidth]{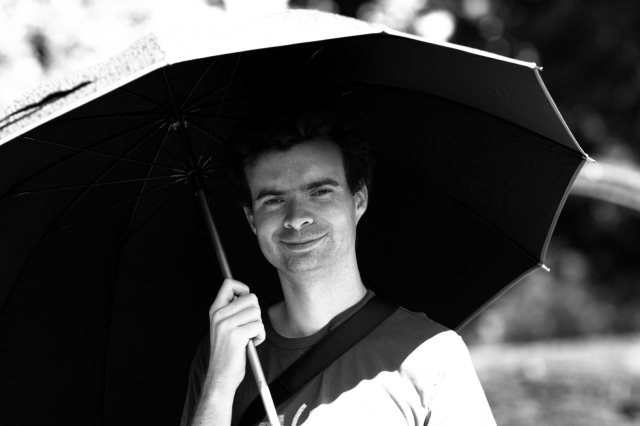}
    \caption{Input}
    \label{fig:face:input}
    \vspace{-4mm}
  \end{subfigure}
  \begin{subfigure}[t]{0.19\linewidth}
    \centering
    \includegraphics[width=\linewidth]{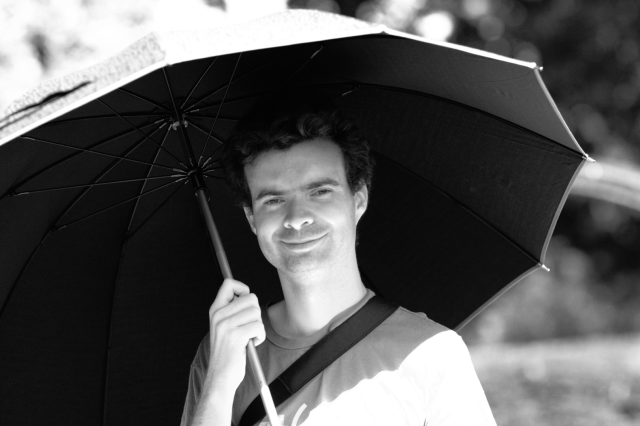}
  \end{subfigure}
  \begin{subfigure}[t]{0.19\linewidth}
    \centering
    \includegraphics[width=\linewidth]{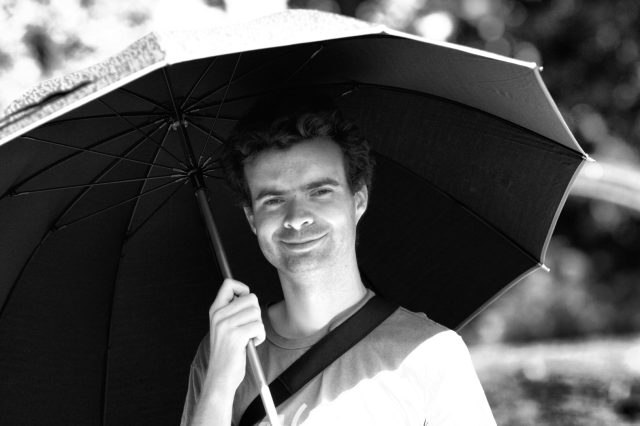}
  \end{subfigure}
  \begin{subfigure}[t]{0.19\linewidth}
    \centering
    \includegraphics[width=\linewidth]{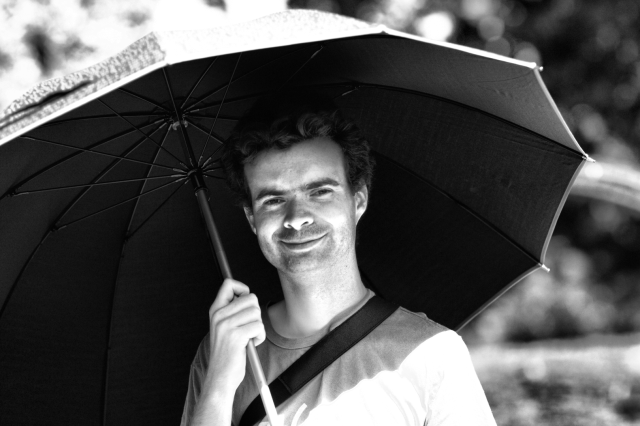}
  \end{subfigure}
  \begin{subfigure}[t]{0.19\linewidth}
    \centering
    \includegraphics[width=\linewidth]{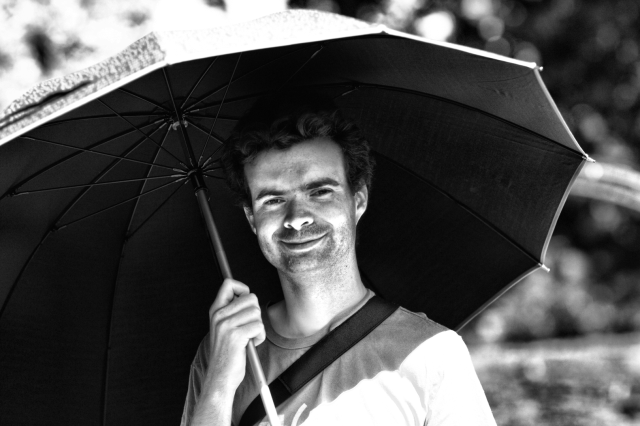}
  \end{subfigure}
  %
  %
  \begin{subfigure}[t]{0.19\linewidth}
    \centering
    \phantom{\includegraphics[width=\linewidth]{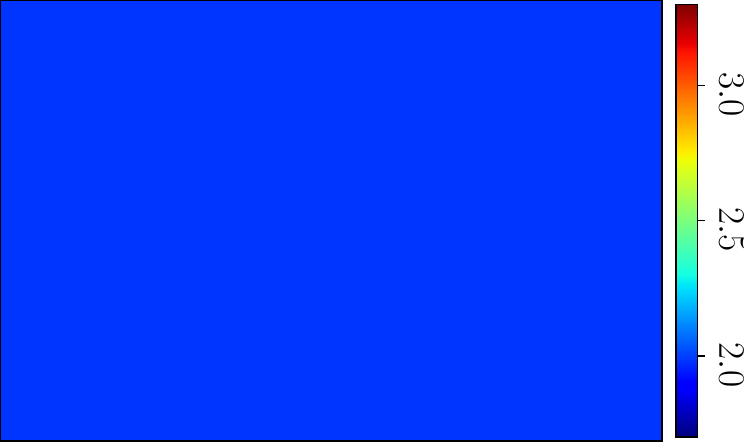}}
  \end{subfigure}
  \begin{subfigure}[t]{0.19\linewidth}
    \centering
    \includegraphics[width=\linewidth]{imgs4publish/ablation/cliplimit_heatmap/a1031-IMG_5026_cliplimit_tgs1.pdf}
    \caption{IA-CLAHE ($1\times 1$)}
    \label{fig:face:1x1}
  \end{subfigure}
  \begin{subfigure}[t]{0.19\linewidth}
    \centering
    \includegraphics[width=\linewidth]{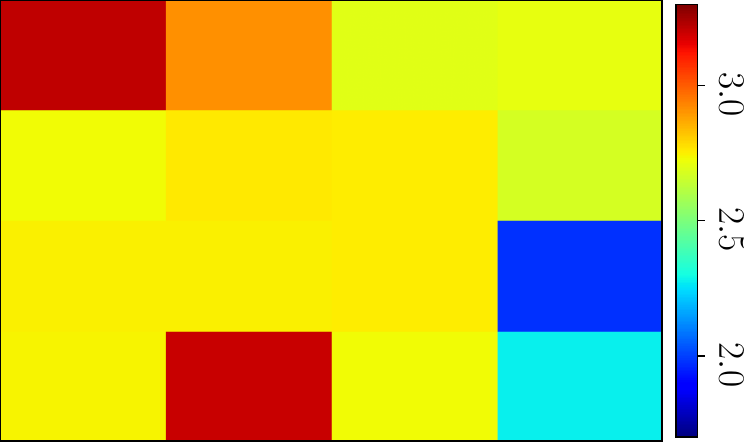}
    \caption{IA-CLAHE ($4\times 4$)}
    \label{fig:face:4x4}
  \end{subfigure}
  \begin{subfigure}[t]{0.19\linewidth}
    \centering
    \includegraphics[width=\linewidth]{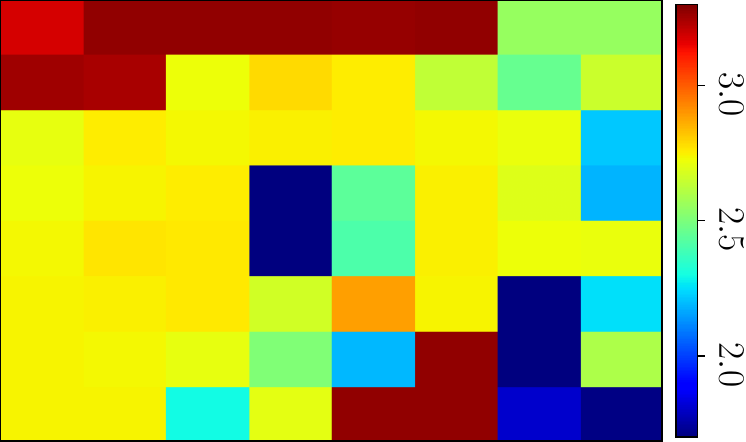}
    \caption{IA-CLAHE ($8\times 8$)}
    \label{fig:face:8x8}
  \end{subfigure}
  \begin{subfigure}[t]{0.19\linewidth}
    \centering
    \includegraphics[width=\linewidth]{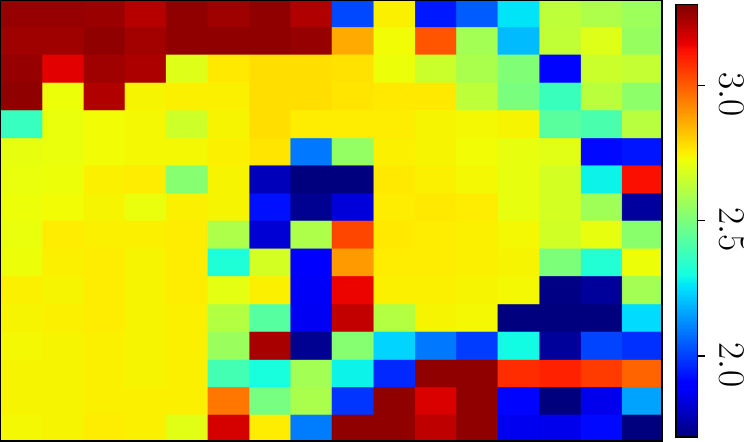}
    \caption{IA-CLAHE ($16\times 16$)}
    \label{fig:face:16x16}
  \end{subfigure}
  \caption{Ablation study on the selection of tile grid size. For scenes with non-uniform luminance distributions (top row), reducing the tile grid size causes flare-spreading. In contrast, for scenes where contrast enhancement is unnecessary (bottom row), increasing the tile grid size leads to over-enhancement. IA-CLAHE can adapt to arbitrary tile grid sizes and estimate scene-appropriate clip limits.}
  \label{fig:ablation}
\end{figure*}

\subsection{Comparison Results on Visual Quality}
\label{sec:expVis}
The visual quality results are presented in Table \ref{tab:visual}. IA-CLAHE achieves effective contrast enhancement while avoiding over-enhancement by tile-wise clip limits. Compared to conventional CLAHE-based methods (CLAHE, RB-CLAHE, LB-CLAHE), IA-CLAHE achieves a better balance across all metrics. CLAHE-based methods often suffer from degraded PSNR/SSIM due to excessive local contrast enhancement, which can lead to over-enhancement artifacts (e.g., excessive skin enhancement or noise amplification). In contrast, our IA-CLAHE improves both PSNR/SSIM and BRISQUE/NIQE, suggesting more natural contrast enhancement.

Figure \ref{fig:qualitative_lcdp} shows how IA-CLAHE adaptively estimates tile-wise clip limits on the LCDP dataset. As shown in the tile-wise clip limits, IA-CLAHE assigns higher clip limits to low-contrast regions (e.g., windows, road on the right), effectively enhancing local contrast in these challenging areas. In contrast, conventional CLAHE-based methods (CLAHE, RB-CLAHE, LB-CLAHE) with single global clip limit tend to over-enhance already high-contrast regions (e.g., man's cheek, tractor), resulting in unnatural appearances.

\subsection{Evaluation Results on Computation Time}
\label{sec:expCom}
We also compared the computation time and number of parameters. All methods were measured on a GPU (NVIDIA GeForce RTX 3080). For fair comparison, only PyTorch-based implementations were included. We evaluated the runtime for processing images with resolutions of 4K ($3840\times2160$) and Full HD ($1920\times1080$). Note that transformer and diffusion-based methods internally resize input images due to their architectural constraints. Therefore, their times reflect reduced-resolution processing rather than original 4K and Full HD sizes. The reported values represent the average runtime over 1000 runs. \#Params denotes the number of parameters, indicating model complexity.

Table \ref{tab:time} shows the results of the computation time comparison. Our IA-CLAHE achieves nearly the same processing speed as standard CLAHE, while being significantly more lightweight than other tone-mapping-based methods (ZeroDCE++ and IA-3DLUT), making it well suited for real-time application. In contrast, transformer and diffusion-based methods require significantly larger models and much longer processing times, making them difficult for real-time application. Although IA-CLAHE is not the fastest method, its runtime is very close to the fastest method (IA-3DLUT).

\subsection{Ablation Study on Tile Grid Size Selection}
Figure \ref{fig:ablation} shows that IA-CLAHE can adapt to arbitrary tile grid sizes and estimate scene-appropriate clip limits. The first row shows an input image (Figure \ref{fig:flare:input}) and results with different tile grid sizes (Figures \ref{fig:flare:1x1} to \ref{fig:flare:16x16}) for a nighttime scene with non-uniform luminance. As the tile grid size increases to $16 \times 16$ (Figure \ref{fig:flare:16x16}), IA-CLAHE suppresses the over-enhancement of the flare region and improves visibility. This behavior can be explained by the clip limits in the second row. At high resolution, the clip limits are spatially sparse with high values concentrated in specific tiles. However, when this map is downsampled to $1 \times 1$ tile grid size where a single clip limit is applied to the entire image, the over-enhancement that spreads the flare occurs. In contrast, the third row shows a daylight scene with a human face and an umbrella (Figure \ref{fig:face:input}) and the enhancement results (Figures \ref{fig:face:1x1} to \ref{fig:face:16x16}). In this scene, larger tile grid sizes lead to over-enhancement of the facial region, resulting in an unnatural appearance. A relatively small tile grid size of $4 \times 4$ (Figure \ref{fig:face:4x4}) achieves better visibility. These results show that IA-CLAHE can accurately estimate spatially-adaptive clip limits, and that selecting an appropriate tile grid size enables better results without additional training.
\section{Conclusion}
We introduced image-adaptive contrast limited adaptive histogram equalization (IA-CLAHE), which estimates tile-wise clip limits (the key parameter controlling contrast enhancement in CLAHE) and applies them adaptively to each local region. We showed that CLAHE is almost everywhere differentiable with respect to clip limit parameters, enabling end-to-end training of the clip limits estimator using image-wise L1 loss. IA-CLAHE was trained only on normal-light images and evaluated in zero-shot image enhancement across unseen adverse weather conditions (low-light, fog, rain, sand, and snow). Results showed that images enhanced by IA-CLAHE simultaneously improve human visibility and machine perception (classification and object detection) compared to prior CLAHE-based methods. IA-CLAHE effectively suppresses local over-enhancement, leading to more stable local contrast while preserving global structure. IA-CLAHE introduces only 211 additional parameters compared to conventional CLAHE and operates at nearly the same speed. These properties allow IA-CLAHE to serve as a drop-in replacement for standard CLAHE with minimal computational overhead, providing an effective and practical solution for robust image enhancement.
{
    \small
    \bibliographystyle{ieeenat_fullname}
    \bibliography{main}
}
\clearpage

\twocolumn[{%
\renewcommand\twocolumn[1][]{#1}%
\maketitlesupplementary
\centering
%
%
\raisebox{-1.8mm}{%
\begin{minipage}[b]{0.24\textwidth}
    \centering
    \includegraphics[width=\linewidth]{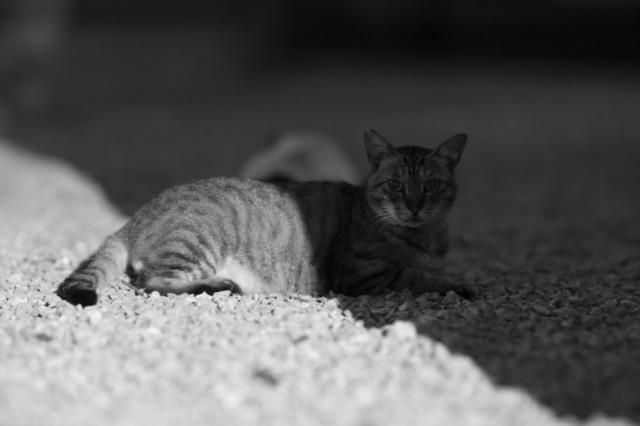}
    \captionof{figure}*{\small (a) Input}
    \vspace{-5.5mm}
\end{minipage}
}%
\hfill
\begin{minipage}[b]{0.24\textwidth}
    \centering
    \includegraphics[width=\linewidth]{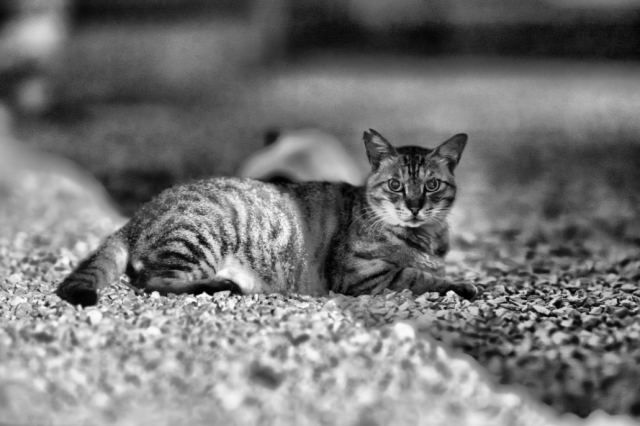}
\end{minipage}
\hfill
\begin{minipage}[b]{0.24\textwidth}
    \centering
    \includegraphics[width=\linewidth]{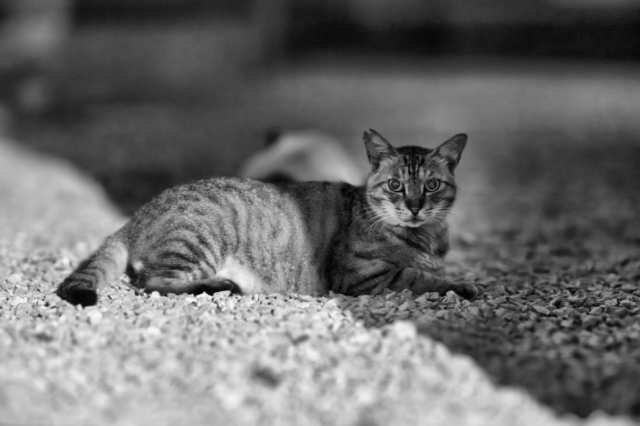}
\end{minipage}
\hfill
\begin{minipage}[b]{0.24\textwidth}
    \centering
    \includegraphics[width=\linewidth]{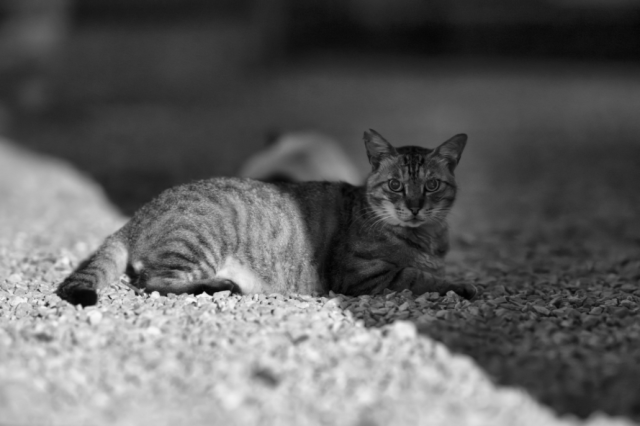}
\end{minipage}
%
%
\begin{minipage}[b]{0.24\textwidth}
    \centering
    \phantom{\includegraphics[width=\linewidth]{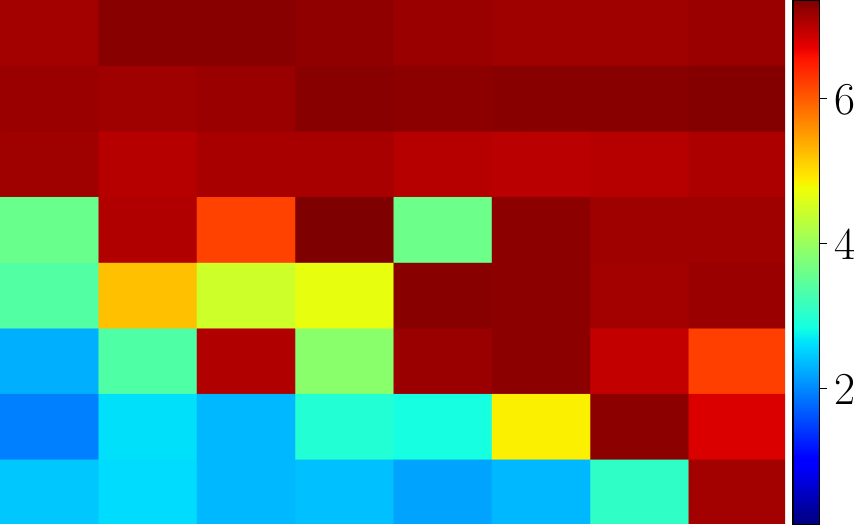}}
\end{minipage}
\hfill
\begin{minipage}[b]{0.24\textwidth}
    \centering
    \includegraphics[width=\linewidth]{imgs4publish/supp/l1_clip_limit_map.pdf}
\end{minipage}
\hfill
\begin{minipage}[b]{0.24\textwidth}
    \centering
    \includegraphics[width=\linewidth]{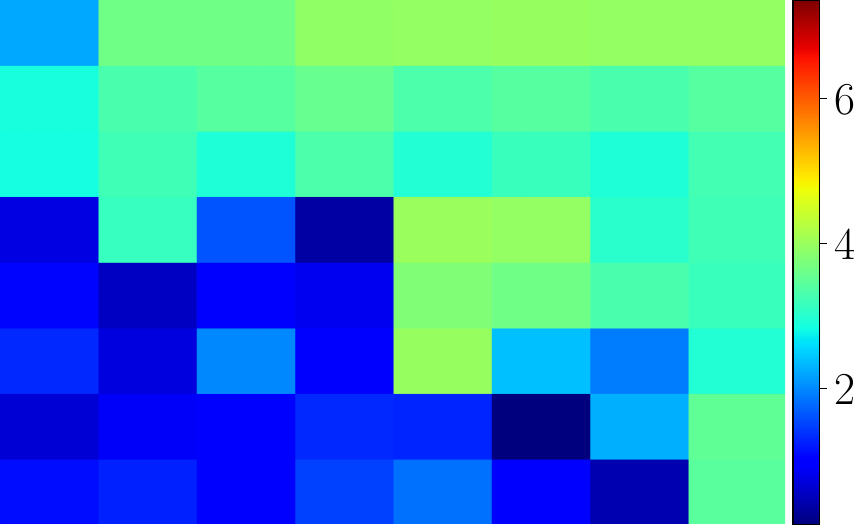}
\end{minipage}
\hfill
\begin{minipage}[b]{0.24\textwidth}
    \centering
    \includegraphics[width=\linewidth]{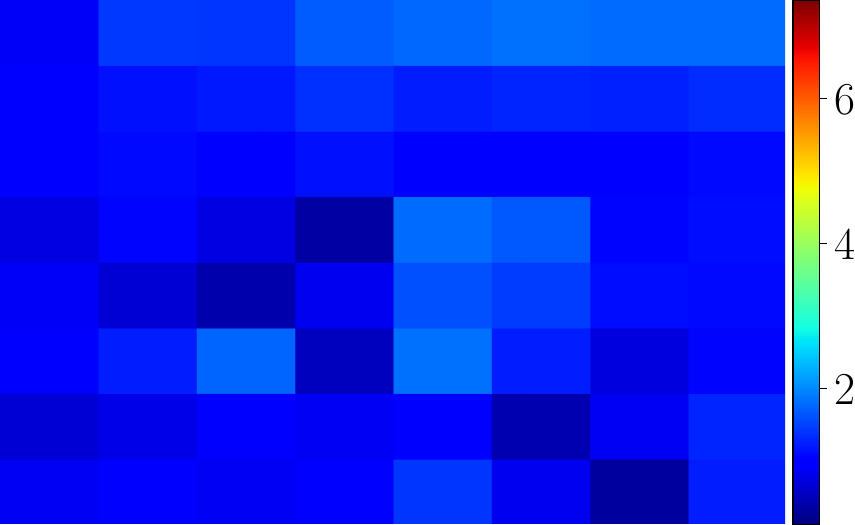}
\end{minipage}
%
%
\raisebox{14mm}{%
\begin{minipage}[b]{0.24\textwidth}
    \vspace{-19mm}
    \centering
    \includegraphics[width=\linewidth]{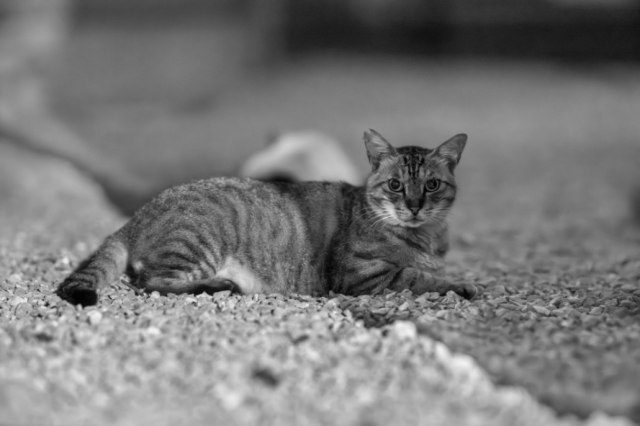}
    \captionof{figure}*{\small (b) Ground-truth}
\end{minipage}
}
\hfill
\begin{minipage}[b]{0.24\textwidth}
    \centering
    \includegraphics[width=\linewidth]{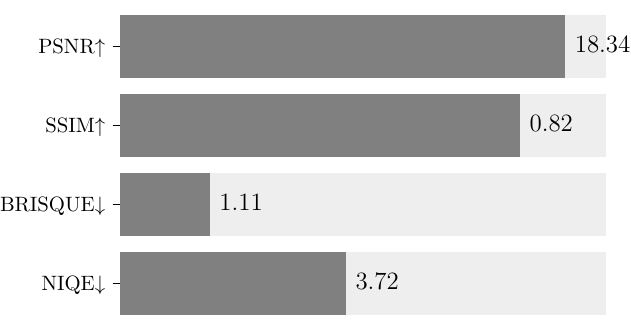}
    \captionof{figure}*{\small (c) L1 loss only}
\end{minipage}
\hfill
\begin{minipage}[b]{0.24\textwidth}
    \centering
    \includegraphics[width=\linewidth]{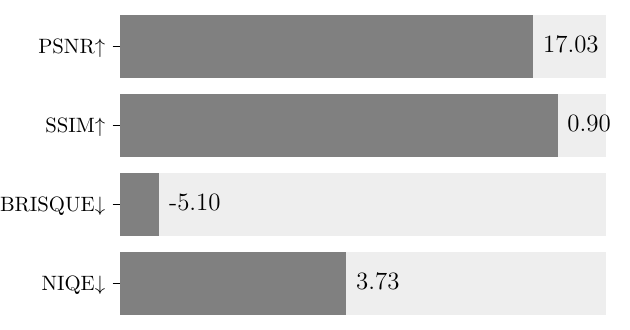}
    \captionof{figure}*{\small (d) SSIM loss only}
\end{minipage}
\hfill
\begin{minipage}[b]{0.24\textwidth}
    \centering
    \includegraphics[width=\linewidth]{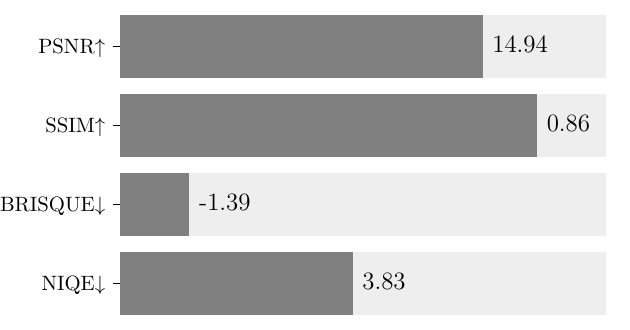}
    \captionof{figure}*{\small (e) Perceptual loss only}
\end{minipage}
\captionof{figure}{Comparison of enhancement results using different loss functions. (a) Input low-light image. (b) Ground-truth. (c-e) Results for L1 loss, SSIM loss, and perceptual loss, respectively. For each loss function, the top row shows the enhanced Y channel in the YCbCr color space, the middle row shows the estimated tile-wise clip limits, and the bottom row shows image quality assessments (IQAs). We evaluated IA-CLAHE using tile grid size of $8\times 8$ to compare tile-wise clip limits across different loss functions.}
\label{fig:loss_comparison}
\vspace{1.0em}
}]
\appendix

This supplementary material presents additional experimental results to support the main paper. Section \ref{sec:supLoss} evaluates our IA-CLAHE framework trained with alternative loss functions (SSIM loss and perceptual loss), demonstrating its flexibility to different optimization objectives. In Section \ref{sec:adaptiveClim}, by comparing with a global clip limit baseline, we demonstrate that tile-wise clip limits are effective for robust enhancement under spatially non-uniform illumination, which is commonly observed in adverse weather conditions.

\section{Comparison with Alternative Objectives}
\label{sec:supLoss}
In the main paper, we adopted the L1 loss, which is widely used in image enhancement literature \cite{Ziwen2024_Real,sun2024_restoring}, to focus on the effectiveness of the proposed method while keeping other factors constant. However, a key advantage of our IA-CLAHE framework is its flexibility to be trained with arbitrary loss functions. In this section, we evaluate the performance of our IA-CLAHE when trained with perceptual losses, demonstrating that our differentiable formulation enables direct end-to-end training with various objectives.

We trained our IA-CLAHE from scratch with three different loss functions individually: L1 loss $L_1$, SSIM loss $L_{ssim}$ \cite{Wang2004SSIM}, and perceptual loss based on VGG-16 \cite{simonyan2014_very} features $L_{per}$ \cite{Johnson2016perceptual}. Training settings followed Section \ref{sec:trainingIACLAHE} with only the loss function modified. To compare tile-wise clip limits across different loss functions, we evaluated IA-CLAHE using tile grid size of $8\times 8$.

\begin{table}[t]
    \centering
    \small
    \caption{Quantitative comparison of different loss functions on the MSEC validation dataset. The metrics are measured on the Y channel in the YCbCr color space. we evaluated IA-CLAHE using tile grid size of $8\times 8$ to compare tile-wise clip limits.}
    \label{tab:perceptual}
    \begin{tabular}{lcccc}
        \hline
        Loss function & PSNR & SSIM & BRISQUE & NIQE \\
        \hline
        $L_{1}$        & 17.39 & 0.79 & \textbf{25.85} & \textbf{3.25} \\
        $L_{\mathrm{ssim}}$ & \textbf{17.48} & \textbf{0.81} & 26.43 & 3.35 \\
        $L_{\mathrm{per}}$  & 17.27 & 0.80 & 27.72 & 3.51 \\
        \hline
    \end{tabular}
\end{table}

\begin{figure*}[t]
\centering
\begin{subfigure}[b]{0.24\textwidth}
    \centering
    \includegraphics[width=\linewidth]{imgs4publish/supp/input_y.png}
\end{subfigure}
\hfill
\begin{subfigure}[b]{0.24\textwidth}
    \centering
    \includegraphics[width=\linewidth]{imgs4publish/supp/gt_y.png}
\end{subfigure}
\hfill
\begin{subfigure}[b]{0.24\textwidth}
    \centering
    \includegraphics[width=\linewidth]{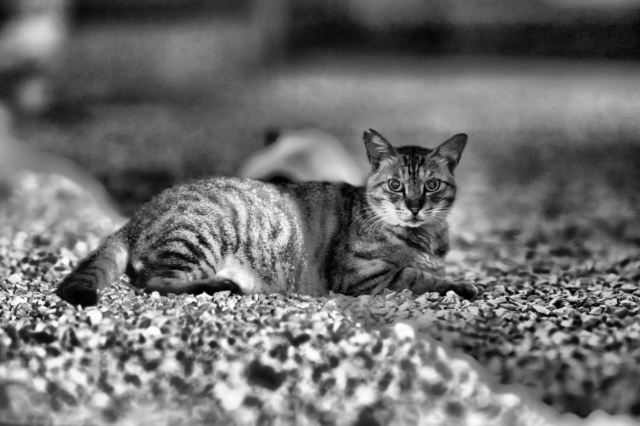}
\end{subfigure}
\hfill
\begin{subfigure}[b]{0.24\textwidth}
    \centering
    \includegraphics[width=\linewidth]{imgs4publish/supp/l1_enhanced_y.png}
\end{subfigure}
%
%
\begin{subfigure}[b]{0.24\textwidth}
    \centering
    \includegraphics[width=\linewidth]{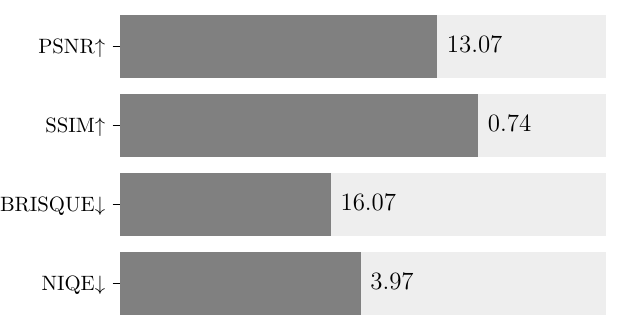}
\end{subfigure}
\hfill
\begin{subfigure}[b]{0.24\textwidth}
    \centering
    \includegraphics[width=\linewidth]{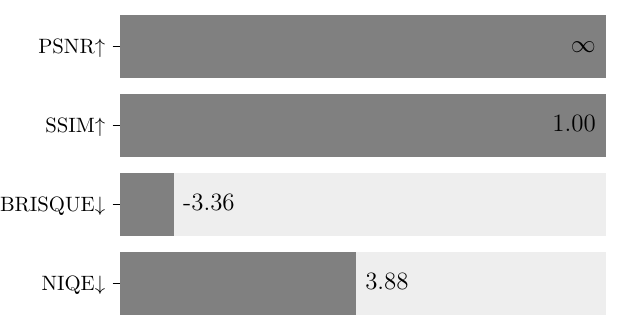}
\end{subfigure}
\hfill
\begin{subfigure}[b]{0.24\textwidth}
    \centering
    \includegraphics[width=\linewidth]{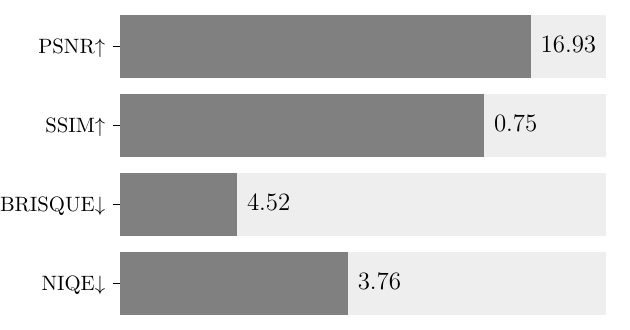}
\end{subfigure}
\hfill
\begin{subfigure}[b]{0.24\textwidth}
    \centering
    \includegraphics[width=\linewidth]{imgs4publish/supp/l1_metric_bars.pdf}
\end{subfigure}
%
%
\begin{subfigure}[b]{0.24\textwidth}
    \centering
    \phantom{\includegraphics[width=\linewidth]{imgs4publish/supp/l1_clip_limit_map.pdf}}
    \caption{Input low-light image}
    \label{fig:tilewise:intput}
\end{subfigure}
\hfill
\begin{subfigure}[b]{0.24\textwidth}
    \centering
    \phantom{\includegraphics[width=\linewidth]{imgs4publish/supp/perceptual_clip_limit_map.pdf}}
    \caption{Ground-truth}
    \label{fig:tilewise:gt}
\end{subfigure}
\hfill
\begin{subfigure}[b]{0.24\textwidth}
    \centering
    \includegraphics[width=\linewidth]{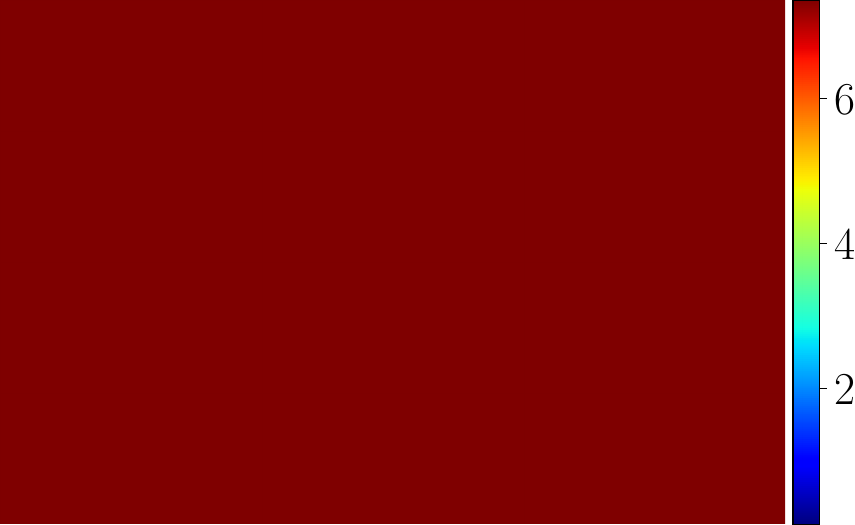}
    \caption{Global clip limit}
    \label{fig:tilewise:global}
\end{subfigure}
\hfill
\begin{subfigure}[b]{0.24\textwidth}
    \centering
    \includegraphics[width=\linewidth]{imgs4publish/supp/l1_clip_limit_map.pdf}
    \caption{Tile-wise clip limits}
    \label{fig:tilewise:local}
\end{subfigure}
\caption{Comparison of global and tile-wise clip limits. (a) Input low-light image. (b) Ground-truth. (c) Enhancement with global clip limit. (d) Enhancement with our adaptive tile-wise clip limits. Middle row: IQAs. Bottom row: clip limit maps ($8 \times 8$ tiles). The global clip limit in (c) is set to the maximum value ($= 7.36$) of the tile-wise clip limits.}
\label{fig:tilewise}
\end{figure*}

Table \ref{tab:perceptual} presents the results on the MSEC validation dataset \cite{Afifi21_msec} when using each loss function individually. L1 loss improves in non-reference metrics (BRISQUE \cite{2012BRISQUE}, NIQE \cite{2013NIQE}) while SSIM loss improves not only non-reference metrics but also full-reference metrics (PSNR, SSIM). This difference reflects the distinct optimization objectives of each loss. L1 loss focuses on minimizing pixel-wise intensity differences, which enables strong contrast enhancement and improves no-reference image quality assessments (IQAs) such as BRISQUE and NIQE. However, this often leads to over-enhancement, resulting in lower PSNR and SSIM compared to SSIM loss. In contrast, SSIM loss optimizes for structural similarity, which naturally balances enhancement and fidelity by preserving both local structures and global consistency. Perceptual loss shows slightly lower performance overall, likely because VGG features are robust to intensity differences and may not strongly penalize contrast variations that deviate from the ground-truth. 

To further analyze how each loss function affects the estimated clip limits, we trained IA-CLAHE on a single image until convergence. Figure \ref{fig:loss_comparison} shows the results, with (a) input image, (b) ground-truth, (c-e) enhanced images (top row), estimated clip limit maps (middle row), and IQAs (bottom row) for each loss function. L1 loss produces strong local contrast enhancement with spatially varying and higher clip limit values, particularly in background regions. SSIM loss yields results structurally closer to the ground-truth with smoother clip limits that preserve structural consistency, resulting in the highest PSNR and SSIM. Perceptual loss generates more conservative enhancement with lower clip limits, as VGG features are less sensitive to intensity variations.

\section{Effectiveness of Adaptive Tile-Wise Clip Limits}
\label{sec:adaptiveClim}
Conventional CLAHE methods apply a single global clip limit across all tiles, which is suboptimal for images with spatially non-uniform illumination, which is commonly observed in adverse weather conditions. Figure \ref{fig:tilewise} shows (a) input low-light image, (b) ground-truth, (c) enhanced result with a global clip limit, and (d) enhanced result with our tile-wise clip limits. The middle row displays the image quality metrics, and the bottom row visualizes the estimated clip limit maps ($8 \times 8$ tiles). The global clip limit in (c) is set to the maximum value ($= 7.36$) of the tile-wise clip limits.

The global clip limit approach (Figure \ref{fig:tilewise:global}) reveals a fundamental trade-off: while it effectively enhances dark regions like the background, it causes severe over-enhancement in brighter areas such as the lower-left foreground, introducing noise amplification and unnatural appearance. Our tile-wise clip limits (Figure \ref{fig:tilewise:local}) resolves this issue by spatially varying clip limits. The estimated clip limit map shows higher values for darker tiles requiring strong enhancement and lower values for brighter tiles requiring conservative enhancement. These results validate that tile-wise adaptation is essential for robust enhancement under spatially non-uniform illumination, which is commonly observed in adverse weather conditions.

\end{document}